\documentclass[11pt,twocolumn]{article}
\usepackage[margin=0.75in]{geometry}
\usepackage{graphicx}
\usepackage{amssymb}
\usepackage{amsmath}
\usepackage{amsfonts}
\usepackage{booktabs}
\usepackage{multirow}
\usepackage{array}
\usepackage{xcolor}
\usepackage{float}
\usepackage{xurl}
\usepackage{textcomp}
\usepackage{hyperref}
\hypersetup{hidelinks}

\newenvironment{tableorg}[1][htbp]{\begin{table}[#1]}{\end{table}}

\begin{document}

% ---------------------------------------------------------------

% ---------------------------------------------------------------
\twocolumn[{%
\begin{center}
{\LARGE \textbf{Denoising-Aware Temporal Point Cloud Completion for 3D Crop Architecture Recovery and Phenotypic Trait Extraction}}\vspace{1.2em}

{\large Mrudul Mittal$^{1}$, Soumyashree Kar$^{2,\ast}$}\vspace{0.6em}

\small $^{1}$Department of Electronics Engineering, Sardar Vallabhbhai National Institute of Technology, Surat, Gujarat, India\\
\small $^{2}$Centre of Studies in Resources Engineering, Indian Institute of Technology Bombay, Mumbai, Maharashtra, India\vspace{0.4em}

\small $^{\ast}$Corresponding author: \texttt{soumyakar@iitb.ac.in} \quad
\texttt{mrudulmittal@gmail.com}
\end{center}

\vspace{1em}
\begin{center}
\parbox{0.92\textwidth}{
\small
\textbf{Abstract.}
High-throughput phenotyping depends on accurate 3D reconstruction of plants across growth stages, yet the development and evaluation of temporal completion methods are limited by the lack of datasets with complete geometric ground truth. To address this challenge, we introduce SynthCrop4D, a procedurally generated synthetic dataset of temporally evolving plant point clouds that provides controllable noise, occlusion, and complete plant geometry for benchmarking reconstruction methods. Using this dataset, we evaluate a two-stage pipeline that combines spatial denoising and temporal point cloud completion. First, a denoising module removes structural artifacts from raw laser-scanned point clouds. The resulting data are then processed by an Adaptive Temporal PoinTr model that reconstructs the current growth stage (t) using information from the previous stage (t--1), enabling recovery of regions missing due to self-occlusion. We evaluate the proposed framework on both SynthCrop4D and the real-world Pheno4D dataset (tomato and maize) under settings with and without denoising. Results show that denoising substantially improves reconstruction quality, with the best configuration achieving a Chamfer Distance of 0.0061 on SynthCrop4D (Temporal PoinTr + Mamba-DG) and an F-Score of 0.2080 on Pheno4D (Vanilla PoinTr + Mamba-DG). We further demonstrate the use of completed point clouds for phenotypic trait extraction, including plant height, canopy width, and convex hull volume, obtaining hull-volume MAEs of 0.021 on synthetic data and 0.343 on real data. Together, SynthCrop4D and the proposed pipeline provide a benchmark and methodology for temporal plant reconstruction and high-throughput crop phenotyping.

\vspace{0.6em}
}
\end{center}
\vspace{1.5em}
}]

%=====================================================================
\section{Introduction}
\label{sec:intro}
%=====================================================================

Precision agriculture increasingly relies on three-dimensional
structural models of crops to automate phenotyping measurements
such as leaf area index, biomass estimation, plant height, and
internode spacing. LiDAR and structured-light sensors offer the
spatial resolution required for organ-level crop architecture
analysis, but the raw point clouds they produce are subject to
two fundamental degradations that prevent direct measurement.
The first is \textbf{sensor noise}: high-frequency jitter and
background outliers that distort fine geometric structures,
particularly thin stems, petioles, and leaf margins, that are
morphologically critical to agricultural phenotyping. The second
is \textbf{self-occlusion}: overlapping leaves in dense crop
canopies create systematic blind spots in which entire leaf
surfaces, axillary structures, and lower internodes are simply
absent from the scan. Unlike general-purpose 3D object datasets
where objects are scanned in isolation, agricultural field scans
combine both degradations simultaneously and at scale. Recent
reviews of 3D crop phenotyping technologies~\cite{li2025crop3drecon}
confirm that this combination of noise and occlusion remains one
of the principal bottlenecks limiting field-scale deployment of
point-cloud-based trait extraction.

Existing deep learning approaches to point cloud
denoising~\cite{rakotosaona2020pointcleannet,luo2021score} were
developed on synthetic or architectural datasets and lack the
crop-specific inductive biases necessary for agricultural
geometry: phytomeric periodicity, bilateral leaf symmetry, and
the strong stem-to-leaf-insertion topology that characterizes
cereal and vegetable crops. Point cloud completion
methods~\cite{yuan2018pcn,yu2021pointr,pang2022pointmae} treat
each scan independently, discarding the longitudinal temporal
information that is uniquely available in phenotyping datasets
where the same plant is scanned at regular growth intervals.
Moreover, denoising and completion are almost universally
evaluated as disjoint tasks, leaving open the question of
whether denoising quality itself affects downstream
reconstruction, and whether the choice of denoiser interacts
with crop morphology in ways that a single fixed pre-processing
pipeline cannot capture.

This paper addresses these gaps with a unified pipeline
evaluated on the Pheno4D dataset~\cite{pheno4d}
(laser-scanned tomato and maize across multiple temporal growth
stages) and a procedurally generated 4D synthetic counterpart,
SynthCrop4D, built specifically for this study (20 maize and
tomato plants across six growth stages, with parametric stem,
leaf, and soil geometry and controlled Gaussian noise
$\sigma{=}0.01$). The contributions of this work are:
\begin{enumerate}
\item A procedurally generated \textbf{4D synthetic crop benchmark},
SynthCrop4D, addressing a key limitation of Pheno4D: real laser
scans have no access to true, noise-free geometry, so real data
alone cannot isolate the effect of sensor noise from that of
self-occlusion. SynthCrop4D provides 20 maize and tomato plants
across six growth stages with paired clean/noisy
($\sigma{=}0.01$) point clouds and controllable occlusion,
enabling supervised denoising training and controlled ablation
that real data cannot support.
\item A \textbf{Mamba-DG} denoising architecture that replaces
the local PointNet~\cite{qi2017pointnet} encoder in
PDN~\cite{suetme2024pdn} with a Hilbert-curve serialized
selective state space encoder, enabling $\mathcal{O}(n)$ global
context aggregation for gradient-field-based crop point cloud
denoising.
\item A baseline study demonstrating that a dense-decoder
temporal network conditioned on the growth stage at $t{-}1$
\textbf{collapses} (Test CD $= 0.287$), performing worse than
a non-temporal self-attention baseline (CD $= 0.150$),
which motivates the design of \textbf{Adaptive Temporal PoinTr}:
a proxy-token completion network that fuses the current partial
scan with the $t{-}1$ prior via temporal cross-attention.

\item An ablation study of a \textbf{pre-processing and completion}
combination between four different denoisers and three different
completion architectures, both for the realistic Pheno4D dataset
and SynthCrop4D. In this test, we see that completing without any
pre-processing leads to a significant degradation no matter what
architecture is used: PCN Baseline has very low F-Score 
($0.007$--$0.009$), as it has no capability of hallucinating 
any additional geometry outside the support of the partial mesh,
while for Temporal PoinTr Test CD score goes from $0.0066$
to $0.0180$ when removing the pre-processing step. We observe
that the optimal denoiser for completion is domain-specific,
as GCN works best for the realistic laser-triangulation noise,
while Mamba-DG is better for the synthetic noise.
\item A \textbf{biological trait analysis} extracting
plant height, canopy width, and convex hull volume from
completed point clouds across growth stages, evaluated on 4
held-out Pheno4D plants (60 stage observations) and 3 synthetic
plants (15 stage observations). Trait error is substantially
lower on synthetic data than on real data (hull-volume MAE of
$0.021$ vs.\ $0.343$), indicating that a large share of
real-world trait error stems from sensor noise and domain gap
rather than from the completion architecture itself.
\end{enumerate}

The remainder of this paper is organized as follows.
Section~2 reviews related work. Section~3 presents the
methodology. Section~4 describes the experimental setup.
Section~5 reports results and discussion. Section~6 concludes.

%=====================================================================
\section{Related Work}
\label{sec:related}
%=====================================================================

\subsection{Point Cloud Denoising}

The approaches for point cloud denoising may be broadly classified as:
displacement regression or score gradient field. PointCleanNet~\cite{rakotosaona2020pointcleannet} can be considered an
example of displacement regression: it employs a PointNet-type~\cite{qi2017pointnet} shared MLP to extract features from local
patch in a PCA normalized local frame and regress directly a 3D displacement vector per point. Such method is very efficient but
local and assumes patch-wise statistical stationarity that does not work in the case of plant architecture.
ScoreDenoise~\cite{luo2021score} introduced score-based
denoising for point clouds, building on the noise-conditional
score network formulation of~\cite{song2019ncsn} to learn the
gradient of the log-probability density of clean points and
applying Langevin dynamics at inference time. PDN~\cite{suetme2024pdn}
extends this to plant-specific geometry by incorporating an
Umbrella Operator Feature (UOF) as a boundary condition that
prevents the gradient ascent from expanding holes in occluded
crop regions.

Graph-based methods such as DGCNN~\cite{wang2019dgcnn} use
dynamic EdgeConv operations on per-layer feature-space $k$-NN
graphs, enabling the network to bridge local geometry and global
plant topology through progressive neighborhood recomputation.

\subsection{Point Cloud Completion}

Point cloud completion aims to reconstruct missing geometry from
a partial scan. PCN~\cite{yuan2018pcn} established the
encoder-decoder paradigm with a coarse-to-fine two-stage decoder
using FoldingNet~\cite{yang2018foldingnet}. PoinTr~\cite{yu2021pointr}
improved this by tokenizing the point cloud into geometric proxy
points and applying Transformer~\cite{vaswani2017attention}
attention to predict missing tokens, in a similar spirit to the
point-wise self-attention formulation of Point
Transformer~\cite{zhao2021pointtransformer}. Point-MAE~\cite{pang2022pointmae}
introduced self-supervised masked autoencoding: random patches
are masked, and the network learns to reconstruct them using a
lightweight ViT-based decoder.

These completion methods treat each scan independently. To our
knowledge, prior work has not used the temporal growth sequences
available in longitudinal agricultural phenotyping datasets as a
structural prior for completion.

\subsection{State Space Models for 3D Data}

Mamba~\cite{gu2023mamba} introduced selective state space models
with $\mathcal{O}(n)$ complexity as a Transformer alternative for
sequence modeling. PointMamba~\cite{liang2024pointmamba} adapted
Mamba to point cloud processing by serializing unordered points
via space-filling curves (Hilbert and Z-order) to impose a
canonical 1D ordering without losing spatial locality. This
yields linear-complexity processing of large point clouds while
maintaining global contextual awareness, making it attractive
for dense field-scale agricultural scans.

\subsection{Agricultural 3D Phenotyping}

PlantNet and PSegNet~\cite{huang2022plantnet} introduced
crop-specific segmentation architectures with 3D
Edge-Preserving Sampling and dynamic graph modules, validated
on tomato, sorghum, and tobacco. DUFA-Net~\cite{dufanet2024}
targets maize canopy segmentation with dual uncertainty-aware
FPS. Pheno4D~\cite{pheno4d} is, to our knowledge, the only
publicly available longitudinal LiDAR dataset with temporal
growth sequences for tomato and maize. Broader surveys of 3D
crop reconstruction technology~\cite{li2025crop3drecon} likewise
identify longitudinal, temporally-aware pipelines as
under-explored relative to single-timepoint reconstruction. The
gap in prior work that this paper addresses is the absence of a
denoising and completion pipeline designed for agricultural crop
architecture that makes use of this temporal structure.

\subsection{Procedural Plant Generation and Trait Extraction}

The use of procedural plant modeling approaches allows creating
parametric 3D geometries of plants based on explicit morphological
rules (length of internodes, leaf number, phyllotaxy angle)
and generating growth sequences for which the ground-truth geometries
are explicitly defined, which is not possible using real-world
sensor data.
Surface properties like leaf area can be usually extracted
from point clouds through the use of Poisson surface
reconstruction techniques~\cite{kazhdan2006poisson}, in which an
implicit surface is fitted to the oriented point normals, although the
reconstructed surface is not guaranteed to be watertight when applied
to noisy or incomplete data and, thus, the correctness of the
reconstruction process needs to be validated.
Volumetric properties such as the convex hull volume have been shown
independently to provide efficient and non-destructive proxy measures of
biomass of LiDAR-sensed row crops~\cite{siebers2024convexhull},
providing thus a good measure to validate the reconstruction process
as well as a biomass proxy.
To the best of our knowledge, previous works on point cloud
completion in agriculture have not made use of synthetic
procedurally generated data with known noise levels as we propose here.

\begin{figure*}[t!]
\centering
\includegraphics[width=\textwidth, height=0.5\textheight, keepaspectratio]{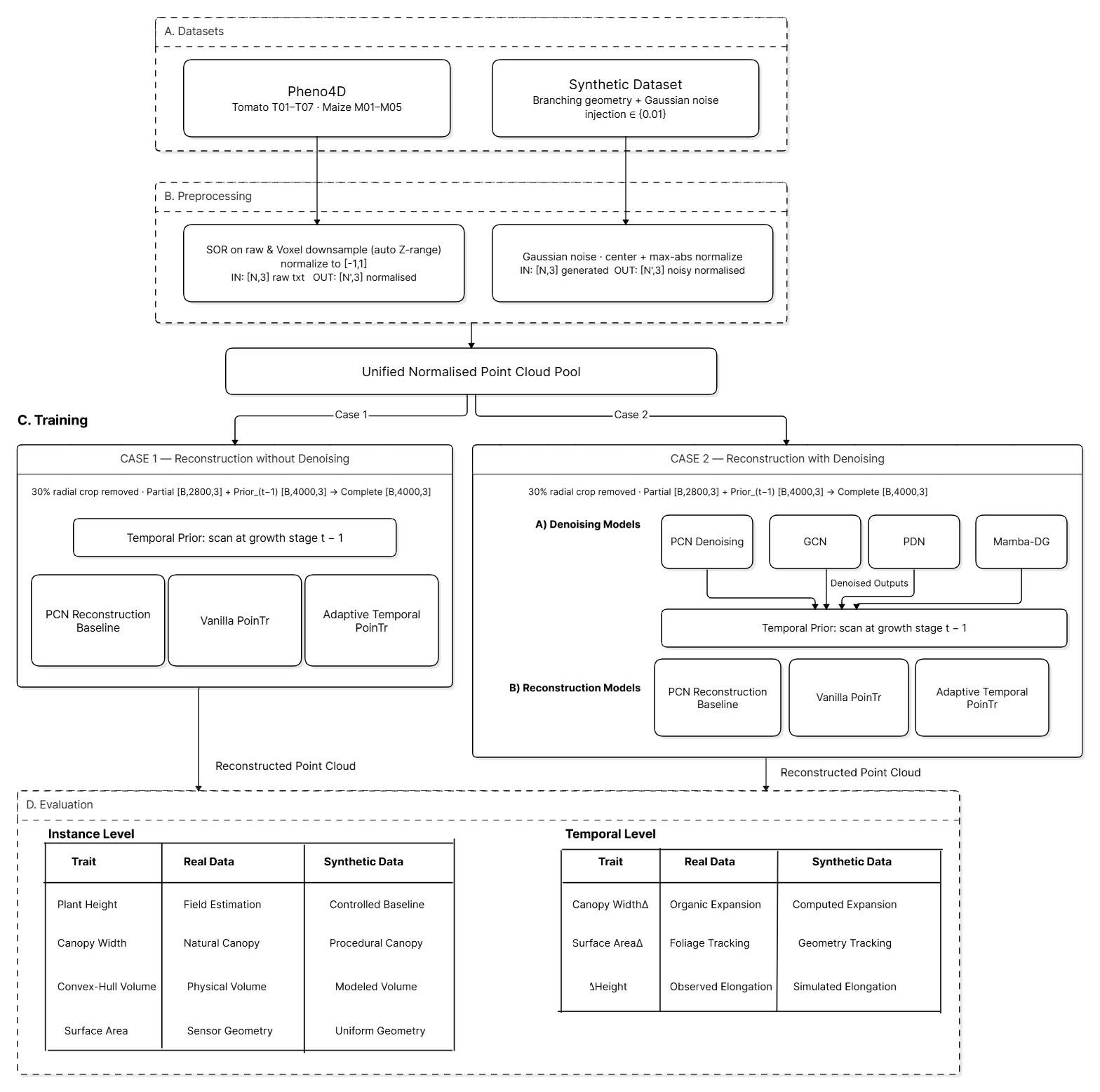}
\caption{Overview of the dual-case evaluation pipeline. Real
(Pheno4D) and synthetic (SynthCrop4D) point clouds are
preprocessed into a unified normalized representation, then
evaluated under two regimes: \textbf{Case~1}, where completion
architectures operate directly on raw, undenoised input, and
\textbf{Case~2}, where one of four denoising models (PCN, GCN,
PDN, Mamba-DG) is applied before completion. Both cases use
the $t{-}1$ growth stage as a temporal prior.}
\label{fig:methodology_pipeline}
\end{figure*}

\subsection{Research Gaps and Objectives}
\label{subsec:gap}

The preceding review highlights several persistent gaps in the
literature that motivate the present work.

\textbf{Lack of domain-specific denoising-completion pipelines.}
Existing point cloud denoising
methods~\cite{rakotosaona2020pointcleannet,luo2021score} are
primarily designed for synthetic object-level benchmarks with
isotropic Gaussian noise. They do not address the complex
artifacts inherent to agricultural structured-light scans, such
as surface-dependent reflectance variations and systematic
canopy occlusions. Furthermore, the impact of denoiser selection
on downstream crop reconstruction remains uninvestigated.

\textbf{Underutilization of temporal growth sequences.}
While longitudinal phenotyping datasets such as
Pheno4D~\cite{pheno4d} provide consecutive growth stages,
current completion
methods~\cite{yuan2018pcn,yu2021pointr,pang2022pointmae} process
scans independently. This discards a robust structural prior,
as occluded regions at stage $t$ are frequently visible in the
prior scan at stage $t{-}1$. No prior work has integrated this
temporal continuity into a completion network via cross-attention.

\textbf{Absence of controlled synthetic agricultural benchmarks.}
Isolating algorithmic reconstruction error from sensor-induced
noise requires precise ground-truth geometry. No paired
clean-and-noisy temporal plant sequences exist for systematic
pipeline ablation in the agricultural domain.

\textbf{Reliance on purely geometric validation.}
Prior completion methodologies are evaluated almost exclusively
on geometric metrics such as Chamfer Distance. Whether
geometrically accurate reconstructions reliably preserve
agronomically critical traits, including plant height, canopy
width, and hull volume~\cite{arend2016phenotyping,siebers2024convexhull},
remains unverified.

\noindent To address these limitations, the objectives of this
paper are to:
\begin{enumerate}
\item Benchmark four denoising architectures on real
agricultural structured-light scans to characterise the
performance ceiling imposed by standard preprocessing filters.
\item Propose Mamba-DG, a Global SSM Gradient Field denoiser
that utilises Hilbert-serialized state space modeling for global
sequential context.
\item Propose Adaptive Temporal PoinTr, a completion network
that explicitly leverages the prior growth stage as a temporal
prior via a cross-attention Transformer decoder.
\item Conduct a systematic 12-configuration ablation
(4 denoisers $\times$ 3 completion models) across both real and
synthetic domains, validating reconstructions against phenotypic
traits and growth dynamics.
\end{enumerate}

%=====================================================================
\section{Methodology}
\label{sec:method}
%=====================================================================

\subsection{Problem Formulation}

Let $\mathcal{P}_t^{noisy} = \{p_i \in \mathbb{R}^3\}_{i=1}^{N}$
denote a noisy, partially occluded point cloud of a crop plant at
growth stage $t$ (a 30\% radial crop simulating self-occlusion),
and $\mathcal{P}_{t-1}^{clean}$ the SOR-filtered scan of the same
plant at the previous growth stage. We define two evaluation
regimes:

\textbf{Case 1 (reconstruction without denoising):}
\begin{equation}
  g_\phi: \left(\mathcal{P}_t^{noisy},\,
  \mathcal{P}_{t-1}^{clean}\right)
  \rightarrow \mathcal{P}_t^{complete}
\end{equation}

\textbf{Case 2 (reconstruction with denoising):}
\begin{equation}
  g_\phi: \left(f_\theta(\mathcal{P}_t^{noisy}),\,
  \mathcal{P}_{t-1}^{clean}\right)
  \rightarrow \mathcal{P}_t^{complete}
\end{equation}

where $f_\theta$ is one of the four Stage-1 denoising models
(Section~\ref{sec:stage1}) and $g_\phi$ is one of the Stage-2
completion architectures (Section~\ref{sec:stage2}). Case~1
corresponds to $f_\theta = \text{identity}$. Figure~\ref{fig:methodology_pipeline}
provides an overview of this dual-case evaluation pipeline.

\subsection{Datasets and Preprocessing}
\label{sec:datasets_preproc}

\subsubsection{Pheno4D Dataset}
\begin{figure*}[t!]
    \centering
    \includegraphics[width=\textwidth, height=0.5\textheight, keepaspectratio]{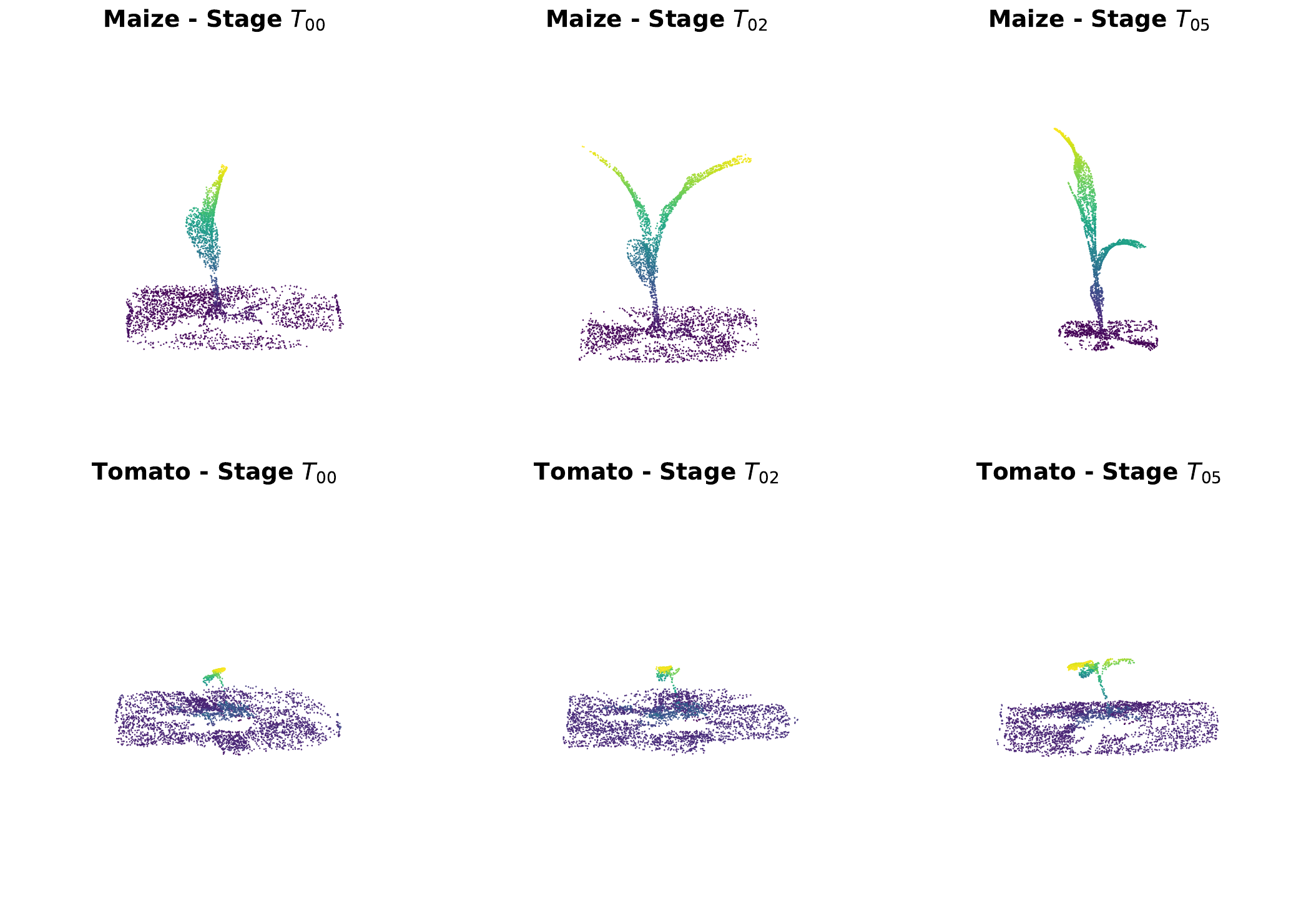}
    \caption{Temporal progression of real 3D plant point clouds from the Pheno4D dataset across different growth stages.}
    \label{fig:temporal_progression}
\end{figure*}

\textbf{Pheno4D}~\cite{pheno4d}: laser-scanned point clouds of
tomato (plants 01--07) and maize (plants 01--05), each captured
at 5--6 temporal growth stages, as illustrated in
Figure~\ref{fig:temporal_progression}. The data was acquired with a
Perceptron ScanWorks V5 laser triangulation scanner mounted on an
articulated measuring arm, which achieves sub-millimetre intrinsic
accuracy ($\sigma \approx 0.012$\,mm at up to 7{,}640 points per scan
line)~\cite{pheno4d}. Consequently, the centimetre-scale residual
irregularity present in the released point clouds predominantly
reflects multi-view registration and stitching artefacts rather than
raw sensor error, which is the noise regime our denoising stage is
designed to target. Per-point organ labels are
available but not used. All coordinates are in metres. We split
strictly by plant identity to prevent temporal leakage:
plants~01--03 of each species for training, plants~04--05 for
validation, and plants~06--07 held out entirely for testing
(74~scans), yielding a three-way train/val/test split. The
74 held-out scans include multiple radially-cropped occlusion
samples generated per growth stage for reconstruction
evaluation. For the biological trait evaluation
(Section~\ref{sec:results}), we further restrict to $T_{06}$,
$T_{07}$, $M_{06}$, $M_{07}$ and use a single observation per
unique growth stage, yielding 60 stage-level trait observations.

\subsubsection{SynthCrop4D: Procedural Synthetic Dataset}
\label{sec:synthcrop}
Pheno4D cannot, by itself, isolate the two degradations studied in
this paper. Its only available clean reference is a Statistical
Outlier Removal (SOR) filter applied to the raw registered scan,
which is itself an estimate rather than exact ground truth, so
denoising error on real data is bounded by filter quality rather than
true geometric error. Moreover, registration noise and self-occlusion
are entangled by construction in a real scan, making it impossible to
vary one while hold the other fixed. To evaluate reconstruction
accuracy and trait-extraction error against an exact ground truth, and
to test whether a denoiser tuned to real, structured noise still
performs well under isotropic synthetic noise, we introduce
\textbf{SynthCrop4D}, a procedural dataset generated using parametric
point-sampling models for maize (\textit{Zea mays}) and tomato
(\textit{Solanum lycopersicum}). This follows the broader paradigm of
functional-structural plant modelling (FSPM), an established approach
for generating 3D crop geometry from explicit architectural
rules~\cite{vos2010fspm}, recently extended to procedural, point-cloud-native
maize reconstruction~\cite{hadadi2025procedural,maizefield3d2025}. Unlike
these prior approaches, which fit procedural surfaces to individual
real LiDAR scans, we generate geometry directly from closed-form
parametric functions, trading per-plant biological fidelity for the
ability to produce paired clean/noisy point clouds with exact,
scan-independent ground truth at scale.

\textbf{Morphology and Parameter Grounding:} We simulate 20 distinct
plants (a random mix of maize and tomato) by sampling a fixed set of
final-stage morphological parameters per plant. These parameters are
grounded in field measurements for mid-vegetative greenhouse stages,
aligning with the experimental context of Pheno4D~\cite{pheno4d}. For
instance, our final synthetic maize heights ($0.60$--$0.90$\,m) target
the upper portion of the V5--V9 vegetative window~\cite{allen2026cornstages}.
Successive maize leaves are distributed around the stem at a fixed
divergence angle of $137.5^\circ$; while mature maize canopies are
empirically closer to a distichous, two-ranked arrangement than to a
golden-angle spiral~\cite{king2004phyllotaxis}, we adopt the golden
angle as a deliberate simplification that guarantees even angular
coverage of the canopy without an explicit two-rank constraint, at the
cost of slightly understating real inter-leaf overlap. Leaf blade
geometry is modelled using parabolic width profiles and quadratic
droop to approximate the natural gravitational deflection of
vegetative maize~\cite{arshad2024maize}.

\textbf{Temporal Growth Model and Splits:} Six growth stages
($t \in \{0, \dots, 5\}$) are generated per plant by linearly scaling
the final-stage height, leaf length, and organ counts from $33\%$ to
$100\%$. While real maize follows a sigmoidal growth
trajectory~\cite{allen2026cornstages}, this linear simplification
creates a controlled structural mismatch. This design isolates
temporal modelling error from sensor noise during the change-point
analysis, as the synthetic inflection point is structurally delayed
compared to the real one.

\begin{figure*}[t!]
    \centering
    \includegraphics[width=\textwidth, height=0.5\textheight, keepaspectratio]{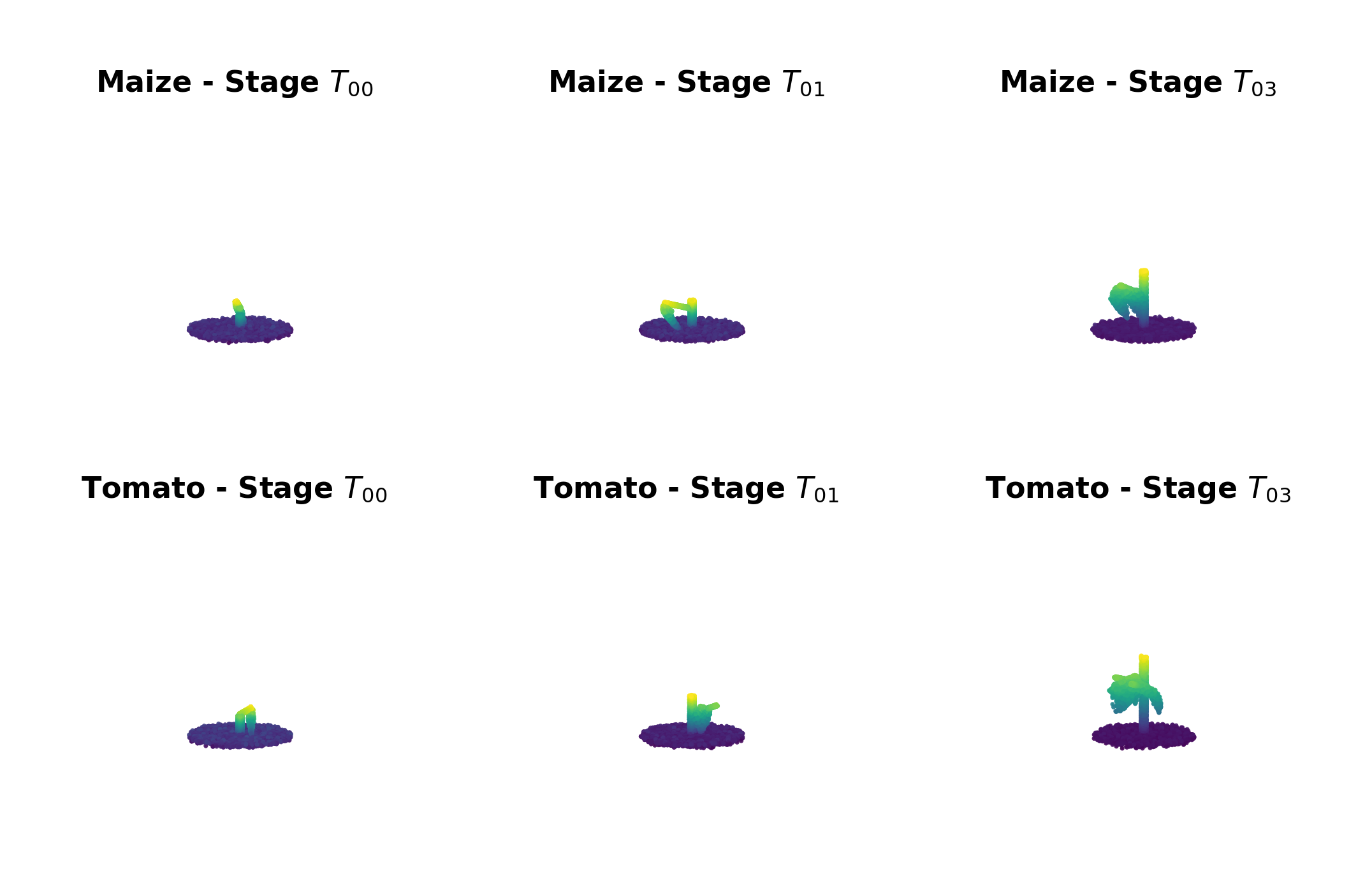}
    \caption{Temporal progression of procedurally generated 3D plant point clouds from the SynthCrop4D dataset across different growth stages.}
    \label{fig:synthcrop_progression}
\end{figure*}

We generate 20 synthetic plant instances $\times$ 6 growth stages $=
120$ paired clean/noisy point clouds. We split by plant index:
plants~00--13 for training (14 plants), plants~14--16 for validation
(3 plants), and plants~17--19 for testing (3 plants). The biological
trait evaluation restricts to the 3 held-out test plants (15
stage-level observations). Because morphological parameters are
sampled independently of Pheno4D plant sizes, SynthCrop4D plants are
on average substantially smaller than real plants (mean hull volume
$0.020$ vs.\ $0.637$\,m$^3$); we account for this scale difference
when interpreting trait error. Each growth stage is rendered to
$N = 4{,}000$ points via uniform surface sampling of the parametric
organs, including a static soil disc. The clouds are then perturbed
with isotropic Gaussian noise ($\sigma = 0.01$) to yield
$t{-}1 \to t$ temporal pairs analogous to real longitudinal sequences.
Visual examples of this synthetic temporal progression for both maize
and tomato are illustrated in Figure~\ref{fig:synthcrop_progression}.
Table~\ref{tab:datasets} summarises the composition, splits, and key
points of correspondence and divergence between both datasets.

\begin{table*}[t!]
\begin{center}
\caption{Correspondence and divergence between Pheno4D and
SynthCrop4D. Rows above the midrule indicate deliberately matched
properties, enabling direct cross-domain comparison; rows below
indicate properties that necessarily differ by construction and
motivate SynthCrop4D's role as a controlled complement to, rather
than a replacement for, real data.}
\label{tab:datasets}
\resizebox{\textwidth}{!}{%
\begin{tabular}{lll}
\toprule
\textbf{Property} & \textbf{Pheno4D (Real)} & \textbf{SynthCrop4D (Synthetic)} \\
\midrule
Species & Maize, tomato & Maize, tomato (matched) \\
Temporal structure & Genuine $t{-}1\!\to\!t$ growth sequences & Genuine $t{-}1\!\to\!t$ growth sequences \\
Growth stages & 5--6 & 6 \\
Points per scan (after preprocessing) & 4{,}000 & 4{,}000 \\
Split protocol & By plant identity (no temporal leakage) & By plant index (no temporal leakage) \\
Perturbation magnitude & Registration residual, $\sim\!\sigma{=}0.01$ scale & Gaussian, $\sigma{=}0.01$ (matched) \\
\midrule
Acquisition modality & Perceptron ScanWorks V5 laser triangulation & Procedural parametric sampling (FSPM) \\
Intrinsic sensor accuracy & $\sigma\approx 0.012$\,mm~\cite{pheno4d} & N/A (exact by construction) \\
Ground truth for denoising & SOR-filtered estimate~\cite{rusu2011pcl} & Exact generative geometry \\
Noise structure & Structured (registration drift, occlusion holes) & Isotropic, i.i.d.\ Gaussian \\
Growth trajectory & Sigmoidal (biological) & Linear ($33\%\!\to\!100\%$, deliberate simplification) \\
Maize phyllotaxy & Distichous, $\sim\!180^\circ$~\cite{king2004phyllotaxis} & Golden angle, $137.5^\circ$ (deliberate simplification) \\
Plant count / scale & 7+7 plants, mean hull volume $0.637\,\mathrm{m}^3$ & 20 plants, mean hull volume $0.020\,\mathrm{m}^3$ \\
Morphological realism & Full biological complexity & Parametric approximation (organ-level, not cellular) \\
\bottomrule
\multicolumn{3}{l}{GT for Pheno4D: SOR-filtered \texttt{\_clean.npy} (nb\_neighbors=20, std\_ratio=2.0).} \\
\multicolumn{3}{l}{GT for SynthCrop4D: exact generative geometry (known per plant/stage).} \\
\end{tabular}%
}
\end{center}
\end{table*}

\subsubsection{Preprocessing}
For Pheno4D, raw scans are processed as follows: (1)~Statistical
Outlier Removal~\cite{rusu2011pcl} on the raw scan
(nb\_neighbors=20, std\_ratio=2.0); (2)~voxel downsampling with
voxel size auto-detected from the Z-range of the first scan in
each sequence; and (3)~normalization to $[-1, 1]$ via center
subtraction and max-absolute scaling, applied consistently to
scan, prior, and ground truth so all three remain in the same
coordinate frame. For SynthCrop4D, point clouds are generated
directly in normalized coordinates; Gaussian noise
($\sigma{=}0.01$) is added post-hoc. All preprocessed clouds
are saved as \texttt{.npy} files to avoid repeating
preprocessing on every epoch.

\subsection{Stage 1: Denoising Pipeline}
\label{sec:stage1}

\subsubsection{PCN: Displacement Regression Baseline}

PCN uses a PointNet-style~\cite{qi2017pointnet} shared MLP with
three Conv1D layers ($3{\to}64{\to}128{\to}256$), global max
pooling to extract a 256-dimensional shape descriptor, and a
per-point displacement regressor that concatenates the global
descriptor with local point coordinates:
\begin{equation}
  \hat{p}_i = p_i + \Delta p_i, \quad
  \Delta p_i = \text{MLP}([p_i \| g(\mathcal{P})])
\end{equation}
Training loss is CD-L1:
\begin{equation}
  \mathcal{L}_{CD} =
    \frac{1}{|\mathcal{P}|}\sum_{p \in \mathcal{P}}
    \min_{q \in \mathcal{Q}} \|p-q\|_1 +
    \frac{1}{|\mathcal{Q}|}\sum_{q \in \mathcal{Q}}
    \min_{p \in \mathcal{P}} \|q-p\|_1
\end{equation}

\subsubsection{GCN: Graph Convolution Baseline}

GCN constructs a $k$-NN graph on the input point cloud and
applies EdgeConv-style graph convolution, computing edge feature
vectors as $(p_i - p_j) \| p_i$ for each neighbor $j$ of point
$i$. Features are aggregated by max pooling within each local
neighborhood. The graph is recomputed after each layer in
feature space, allowing semantic neighbors to replace geometric
neighbors as the representation matures.

\subsubsection{PDN: Local Gradient Field}

PDN~\cite{suetme2024pdn} computes per-point features via a
PointNet-based~\cite{qi2017pointnet} FeatureExtraction module
(60-dimensional) and augments them with an Umbrella Operator
Feature (UOF):
\begin{equation}
  \text{UOF}(p_i) =
    \frac{1}{K}\sum_{j \in \mathcal{N}(p_i)}(p_j - p_i)
\end{equation}
The combined features condition a ScoreNet that predicts
$\nabla_p \log q(p)$, following the noise-conditional score
matching framework of~\cite{song2019ncsn}. Training minimises
the Denoising Score Matching (DSM) objective:
\begin{equation}
\begin{split}
  \mathcal{L}_{DSM} &= \frac{1}{2\sigma}
  \mathbb{E}\left[\left\|
    s_\theta(p + \epsilon) - (-\epsilon)
  \right\|^2\right], \\
  &\quad \epsilon \sim \mathcal{N}(0, \sigma^2 I)
\end{split}
\end{equation}
Inference applies Langevin dynamics for 10 steps with step
size~0.2. PDN was initialised from official pre-trained weights
and fine-tuned on Pheno4D.

\subsubsection{Mamba-DG: Global SSM Gradient Field}

\begin{figure*}[t!]
\centering
\includegraphics[width=\textwidth, height=0.45\textheight, keepaspectratio]{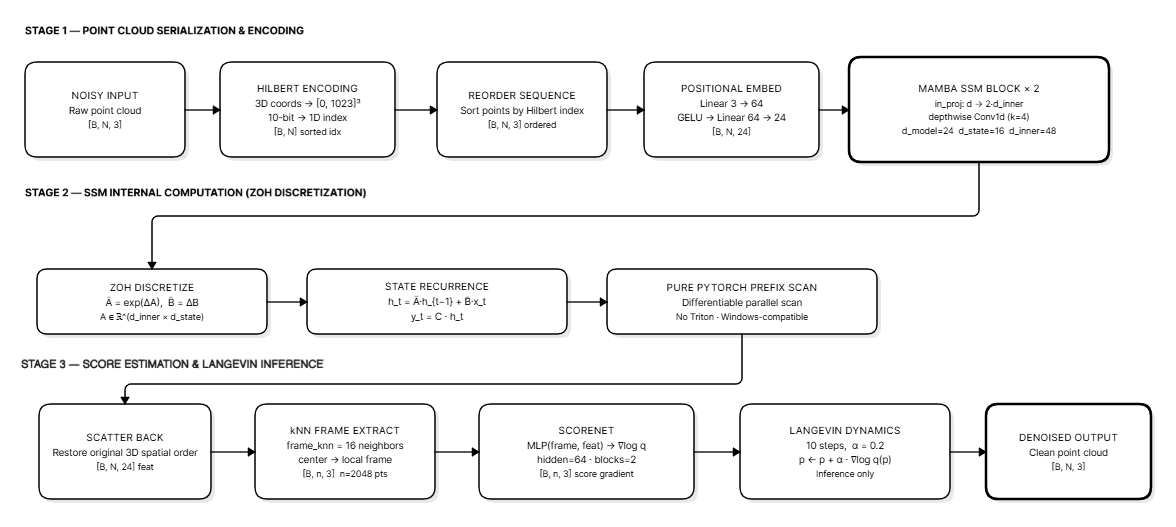}
\caption{Architecture of the proposed Mamba-DG denoiser.
The model operates in three stages: (1)~3D points are
serialized via Hilbert Encoding and processed through Mamba
SSM blocks to capture global context; (2)~SSM computation
uses ZOH discretization and a pure PyTorch parallel prefix
scan; (3)~features are scattered back to 3D space, combined
with local $k$-NN frames, and fed into a ScoreNet followed
by Langevin dynamics for iterative denoising.}
\label{fig:mamba_dg_arch}
\end{figure*}

Mamba-DG replaces PDN's local PointNet encoder with a
PointMambaEncoder that processes points as an ordered 1D
sequence via Hilbert curve serialization. Given raw 3D
coordinates normalised to $[0, 1023]^3$, a 10-bit Hilbert
encoding maps each point to a 1D index preserving 3D spatial
locality. Points are reordered to form a sequence
$\mathbf{x} = (x_1, \ldots, x_N)$ processed by two stacked
Mamba SSM blocks. Each block applies ZOH discretization:
\begin{equation}
  \bar{A} = e^{\Delta A}, \qquad \bar{B} = \Delta B
\end{equation}
and computes the state recurrence:
\begin{equation}
  h_t = \bar{A} h_{t-1} + \bar{B} x_t, \qquad y_t = C h_t
\end{equation}
The recurrence is implemented as a differentiable parallel
prefix scan in pure PyTorch, requiring no Triton CUDA kernels.
After processing, point features are scattered back to the
original 3D order. The ScoreNet DG head is identical to PDN.
Mamba-DG was trained from scratch on Pheno4D.
Figure~\ref{fig:mamba_dg_arch} illustrates the full three-stage
architecture. Table~\ref{tab:denoise_arch} summarises the
mechanism, loss, and key characteristics of all four denoising
architectures compared in this study, while
Table~\ref{tab:complete_arch} provides the corresponding
summary for the three completion architectures introduced in
Section~\ref{sec:stage2}.

\begin{tableorg}[t!]
\centering
\caption{Denoising architectures compared
(Section~\ref{sec:stage1}).}
\label{tab:denoise_arch}
\resizebox{\columnwidth}{!}{%
\begin{tabular}{llcl}
\toprule
\textbf{Model} & \textbf{Mechanism} &
\textbf{Loss} & \textbf{Notes} \\
\midrule
PCN & PointNet$\to$displacement & CD-L1 &
  Local only \\
GCN & EdgeConv on $k$-NN graph & CD-L1 &
  Feature-space graph \\
PDN & PointNet + UOF $\to$ Langevin 
& DSM &
  Pretrained \\
Mamba-DG & Hilbert SSM $\to$ Langevin & DSM &
  $\mathcal{O}(n)$ global \\
\bottomrule
\end{tabular}%
}
\end{tableorg}

\begin{tableorg}[t!]
\centering
\caption{Completion architectures compared
(Section~\ref{sec:stage2}).}
\label{tab:complete_arch}
\resizebox{\columnwidth}{!}{%
\begin{tabular}{llcl}
\toprule
\textbf{Model} & \textbf{Mechanism} &
\textbf{Uses $t{-}1$} & \textbf{Notes} \\
\midrule
PCN & FC decoder on global desc. & No &
  Displaces points \\
Vanilla PoinTr & FPS proxy + self-attn & No &
  Ablation \\
Adaptive PoinTr &
  Self-attn + cross-attn & Yes &
  Ours \\
\bottomrule
\end{tabular}%
}
\end{tableorg}

\subsection{Stage 2: Completion Architectures}
\label{sec:stage2}

\subsubsection{PCN: Displacement-Only Baseline}

As a minimal completion baseline, we adapt PCN to the
completion setting: a PointNet-style~\cite{qi2017pointnet}
encoder reduces the partial input to a 256-dimensional global
descriptor, decoded through a fully connected head into a
fixed-size point set. Because this architecture has no mechanism
for inserting new points beyond the partial input's support, it
can only displace points already present in the partial scan. We
include it as a lower-bound reference, and discuss its
characteristic failure mode (near-zero F-Score despite
competitive Chamfer Distance) in Section~\ref{sec:ablation}.

\subsubsection{Dense-Decoder Temporal Fusion}
\label{sec:pilot}

Our first attempt at temporal completion used a dual-branch
PointNet~\cite{qi2017pointnet} encoder to extract 256-dimensional
global descriptors for both the current partial frame
($f_{partial}$) and the prior clean frame ($f_{prior}$),
concatenated and passed through a fully connected decoder:
\begin{equation}
  \hat{\mathcal{P}}_t =
  \text{Decoder}_{FC}([f_{partial} \| f_{prior}])
\end{equation}
Despite resolving a BatchNorm singularity
(\texttt{drop\_last=True}) and applying standard regularisation
(Gaussian jitter, Z-rotation, weight decay~$0.1$,
Dropout($0.3$)), the model collapsed
(Section~\ref{sec:pilot_results}), performing worse than a
non-temporal self-attention baseline. This motivated the
proxy-token cross-attention architecture below.

\subsubsection{Adaptive Temporal PoinTr}

Adaptive Temporal PoinTr tokenizes both the partial scan
$\mathcal{P}_t$ and the prior scan $\mathcal{P}_{t-1}$ into
proxy points via FPS, producing proxy feature sets
$\mathcal{F}_t$ and $\mathcal{F}_{t-1}$ through a shared
Conv1D encoder. $\mathcal{F}_t$ is refined by a spatial
Transformer~\cite{vaswani2017attention} encoder, then a
Transformer decoder applies cross-attention with
$\mathcal{F}_{t-1}$ as memory (Key/Value) and the
spatially-refined $\mathcal{F}_t$ as Query:
\begin{equation}
  \mathcal{F}_t^{fused} = \text{CrossAttn}(
    Q{=}\text{SelfAttn}(\mathcal{F}_t),\;
    K/V{=}\mathcal{F}_{t-1})
\end{equation}

\begin{figure*}[t!]
\centering
\includegraphics[width=\textwidth, height=0.45\textheight, keepaspectratio]{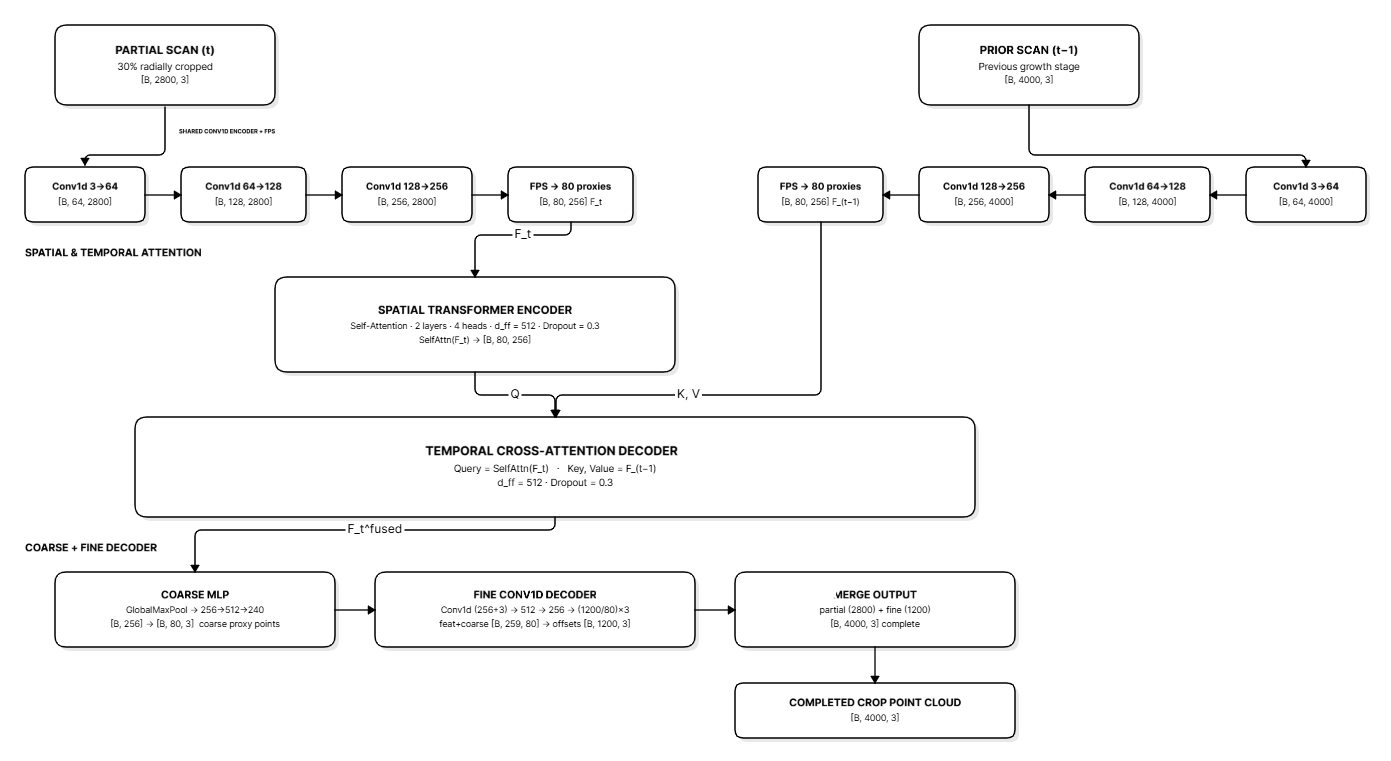}
\caption{Architecture of the Adaptive Temporal PoinTr
completion model. Both the current partial scan at $t$ and the
complete prior scan at $t{-}1$ are encoded and downsampled into
proxy tokens via FPS. A Spatial Transformer refines the current
frame's features; a Temporal Cross-Attention Decoder queries
the prior scan's geometry. A coarse-to-fine MLP decoder then
generates the missing geometry.}
\label{fig:adaptive_temporal_pointr_arch}
\end{figure*}

As shown in Figure~\ref{fig:adaptive_temporal_pointr_arch}, this
allows the network to explicitly query the prior growth
stage's geometry when completing occluded regions at time $t$.
A coarse MLP predicts proxy-level completions from
$\mathcal{F}_t^{fused}$, and a fine Conv1D head refines each
proxy into a local point patch. \textbf{Vanilla PoinTr} is an
ablation variant that omits $\mathcal{F}_{t-1}$ and the
cross-attention decoder entirely.

\subsection{Mesh-Derived Trait Extraction}
\label{sec:traits}

To examine whether reconstructed point clouds preserve
biologically relevant structure, we convert both ground-truth
and reconstructed point clouds to surface meshes via Poisson
surface reconstruction~\cite{kazhdan2006poisson} (depth~8)
after normal estimation and consistent tangent-plane
orientation. From each mesh we extract \textbf{surface area}
(a leaf-area proxy); we additionally record a binary
\textbf{watertightness} flag, since Poisson reconstruction on
partial, noisy agricultural point clouds frequently fails to
close the surface: none of the meshes in our evaluation
(real or synthetic, ground-truth or reconstructed) were
watertight, so surface-area values should be read as
approximate. Directly from the point cloud (mesh-independent)
we compute three primary traits: \textbf{height} (Z-axis
extent), \textbf{canopy width} (XY bounding box diagonal),
and \textbf{hull volume} (convex-hull volume), the latter
following the LiDAR-based convex-hull biomass-proxy
methodology of~\cite{siebers2024convexhull}. For consecutive
growth stages, we compute \textbf{growth deltas} and report
sign agreement between predicted and ground-truth deltas as
a coarser, more robust signal of whether the model tracks
plausible growth direction.

Alongside these primary traits, we compute secondary shape
descriptors: vertical center of mass, projected canopy area,
effective LAI proxy, canopy closure, compactness ratio, and
elongation ratio. These are reported in supplementary results
but are not treated as primary evaluation metrics since
several are derived ratios of the primary traits above.

\subsection{Delta-Conditioning and Trait-Aware
Fine-Tuning}
\label{sec:v2}

Initial evaluation of the baseline Adaptive Temporal PoinTr
pipeline (Section~\ref{sec:stage2}) revealed two systematic
failure modes in trait-level accuracy. First, hull volume
was consistently overestimated because the temporal
cross-attention decoder passively imports $t{-}1$ geometry
without accounting for inter-stage plant growth: the decoder
treats the prior-stage scan as a valid description of the
current stage, inflating the reconstructed convex hull with
stale geometry. Second, the CD-L2 completion loss is
agnostic to macroscopic trait values a network can
achieve low CD by distributing points across a locally
plausible surface without recovering the correct global
height or canopy extent.

We introduce two targeted modifications to address these
failure modes, collectively forming the \textbf{V2 pipeline}.
These modifications apply only to the real-data evaluation;
the synthetic SynthCrop4D experiments use the baseline
architecture throughout.

\noindent\textbf{Delta-conditioning.}
We augment the Adaptive Temporal PoinTr module~\cite{yu2021pointr} with a
\texttt{growth\_estimator} that computes a residual feature
between the current and prior proxy token sets~\cite{qi2017pointnetplusplus}, and an
\texttt{adaptive\_gate} that learns to weight this correction
per proxy token:
\begin{equation}
  \mathcal{F}_{t-1}^{\text{cond}} =
  \mathcal{F}_{t-1} +
  \sigma\!\left(W_g\,\mathcal{F}_t\right)
  \odot
  \left(\mathcal{F}_t - \mathcal{F}_{t-1}\right)
  \label{eq:delta_cond}
\end{equation}
where $\sigma$ is the sigmoid activation and $W_g$ is a
learned projection. This delta-conditioned prior
$\mathcal{F}_{t-1}^{\text{cond}}$ replaces the static
$\mathcal{F}_{t-1}$ as Key/Value in the temporal
cross-attention decoder~\cite{vaswani2017attention},
forcing the decoder to query a growth-adjusted prior
rather than an unchanged snapshot of the previous stage.

\noindent\textbf{Trait-aware fine-tuning.}
We augment the CD-L2 completion loss~\cite{fan2017point} with a lightweight
trait consistency term that directly penalises discrepancies
in macroscopic geometric descriptors:
\begin{equation}
  \mathcal{L}_{\text{V2}} = \mathcal{L}_{\text{CD}} +
  \lambda_h \|\hat{h} - h\|_1 +
  \lambda_w \|\hat{w} - w\|_1
\end{equation}
where $\hat{h}$ and $\hat{w}$ are the predicted height
(Z-axis extent) and canopy width (XY bounding box diagonal),
extracted directly from the output point cloud, and $h$,
$w$ are the corresponding ground-truth values. We use
$\lambda_h = \lambda_w = 0.1$ throughout, chosen to keep
the trait terms at roughly the same scale as the CD loss
without dominating it. The trait terms are computed without
any mesh reconstruction and add negligible overhead to
the training loop.

%=====================================================================
\section{Experimental Setup}
\label{sec:exp}
%=============================================================

\subsection{Implementation Details}

All models are implemented in PyTorch. Denoising models (PCN,
GCN, Mamba-DG) are trained from scratch on Pheno4D; PDN is
initialised from official pre-trained weights and fine-tuned.
Adaptive Temporal PoinTr is initialised by transferring weights
from Vanilla PoinTr (\texttt{transformer.*} $\to$
\texttt{spatial\_transformer.*}) before fine-tuning the temporal
cross-attention decoder. Tables~\ref{tab:hyperparams_denoise}
and~\ref{tab:hyperparams_completion} summarise hyperparameters,
broken out per architecture given the substantial differences
in optimisation regime across paradigms. All experiments use
PyTorch on NVIDIA RTX~3050 (6\,GB), Google Colab T4 (16\,GB),
and Kaggle T4$\times$2 (16\,GB).

\begin{table*}[t!]
\begin{center}
\caption{Denoising-model training hyperparameters. Selected
independently per architecture rather than via a shared grid
search, since the four denoisers differ in loss paradigm,
initialisation, and memory footprint.}
\label{tab:hyperparams_denoise}
\begin{tabular}{lllll}
\toprule
\textbf{Hyperparameter} & \textbf{PCN} & \textbf{GCN} &
\textbf{PDN} & \textbf{Mamba-DG (Ours)} \\
\midrule
Loss                    & CD              & CD              & DSM                        & DSM \\
Optimiser               & AdamW           & AdamW           & AdamW                      & AdamW \\
Learning rate           & $1\times10^{-3}$ & $1\times10^{-3}$ & $5\times10^{-5}$          & $1\times10^{-3}$ \\
Weight decay            & 0               & $1\times10^{-4}$ & 0                          & 0 \\
LR schedule             & Fixed (none)    & Cosine annealing & Cosine annealing         & Cosine annealing \\
Batch size              & 1               & 1               & 1                          & 1 \\
Epochs                  & 30              & 30              & 30                         & 30 \\
Max points (train)      & 6{,}000         & 4{,}000         & 6{,}000                    & 15{,}000 \\
Max points (test)       & 15{,}000        & 15{,}000        & 15{,}000                   & 15{,}000 \\
Gradient clipping       & None            & None            & Yes (max\_norm = 1.0)      & None \\
Dropout                 & --              & None            & --                         & None \\
Initialisation          & Scratch         & Scratch         & Pretrained (fine-tuned)    & Scratch \\
\bottomrule
\end{tabular}
\end{center}
\end{table*}

\begin{table*}[t!]
\begin{center}
\caption{Completion-model training hyperparameters. Unlike the
denoisers, all three completion architectures share an
identical recipe.}
\label{tab:hyperparams_completion}
\begin{tabular}{l p{10cm}}
\toprule
\textbf{Hyperparameter} & \textbf{PCN Baseline / Vanilla PoinTr / Adaptive Temporal PoinTr} \\
\midrule
Loss              & Chamfer Distance (mean NN pred$\to$GT + mean NN GT$\to$pred); no auxiliary term \\
Optimiser         & AdamW \\
Learning rate     & $1\times10^{-4}$ (single shared rate; no backbone/decoder split) \\
Weight decay      & $1\times10^{-4}$ \\
Batch size        & 2 \\
Epochs (max)      & 100 \\
Early stopping    & Patience 20 epochs, monitored on validation CD-L2 \\
Max points        & 4{,}000 total per cloud (2{,}800 observed partial input + 1{,}200 target region) \\
Initialisation    & PCN Baseline, Vanilla PoinTr: scratch. Temporal PoinTr: backbone transferred \\
\bottomrule
\end{tabular}
\end{center}
\end{table*}

\begin{table*}[t!]
\begin{center}
\caption{Denoising results on 74 held-out Pheno4D test clouds (30-epoch checkpoints, 15,000 max points). While PCN and GCN train using CD and PDN/Mamba-DG use DSM, we report unified Chamfer Distances (CD) across all splits to ensure direct cross-comparability.}
\label{tab:denoising_main}
\begin{tabular}{llcccc}
\toprule
\textbf{Model} & \textbf{Paradigm} &
\textbf{Train CD$\downarrow$} & \textbf{Val CD$\downarrow$} &
\textbf{Test CD$\downarrow$} & \textbf{F@5mm$\uparrow$} \\
\midrule
PCN            & Displacement Reg. (CD)  & 0.0922 & 0.0899 & 0.0862 & 0.0453 \\
GCN            & Graph Conv. (CD)        & \textbf{0.0919} & \textbf{0.0896} & 0.0861 & \textbf{0.0472} \\
PDN            & Local DG (DSM)           & 0.0922 & 0.0899 & 0.0863 & 0.0471 \\
\textbf{Mamba-DG} & \textbf{Global SSM + DG} & \textbf{0.0919} & 0.0897 & \textbf{0.0860} & 0.0471 \\
\bottomrule
\end{tabular}
\end{center}
\end{table*}

\subsection{Evaluation Metrics}

\textbf{Chamfer Distance}: bidirectional average
nearest-neighbour distance in normalised space between predicted
and ground-truth point sets, reported as CD-L1
(Sections~\ref{sec:denoising_results}--\ref{sec:pilot_results})
or CD-L2 (Section~\ref{sec:ablation}). Lower is better.

\textbf{F-Score @ 5mm}: fraction of predicted points within
threshold $\tau = 0.01$ of the normalised coordinate range
(${\approx}5$\,mm at real scale), reported as the harmonic
mean of precision and recall. Higher is better. We note that
this metric can be inflated by architectures that largely
reproduce input points rather than generating new geometry,
particularly when the input has not been denoised; we report
it alongside CD rather than in isolation.

\textbf{Inference time}: wall-clock milliseconds per scan on
NVIDIA RTX~3050, averaged over the test set.

\textbf{Trait MAE}: mean absolute error between ground-truth
and reconstructed trait values. Because absolute trait
magnitudes differ substantially between Pheno4D and SynthCrop4D,
we additionally report relative trait MAE (MAE / mean GT).

\textbf{Growth-delta sign agreement}: proportion of consecutive
growth-stage pairs where the predicted change in a trait has
the same sign as the ground-truth change. Invariant to absolute
trait scale, enabling direct comparison across real and
synthetic domains.

\subsection{Ablation Protocol}
\label{sec:ablation_protocol}

To systematically isolate the impact of each pipeline component, we designed a 12-configuration ablation study evaluating all combinations of our four denoising and three completion models across both datasets.

\textbf{Evaluation Regimes.}
We test under two conditions: \textbf{Case 1 (Without Denoising)}, where raw, noisy partial scans are fed directly into the completion network to establish a baseline; and \textbf{Case 2 (With Denoising)}, where scans are first preprocessed by one of the four denoisers before completion (yielding $4 \times 3 = 12$ distinct pipelines).

\textbf{Occlusion and Temporal Priors.}
To simulate self-occlusion, we drop the nearest 30\% of points around a randomly selected center, yielding a partial input $\mathcal{P}_t^{\text{partial}} \in \mathbb{R}^{2800 \times 3}$. We consistently normalise this input, the ground truth, and the temporal prior using the partial scan's center and scale. The temporal prior $\mathcal{P}_{t-1}^{\text{clean}}$ is simply the clean scan of the same plant from the previous growth stage (excluding $t=0$, which inherently lacks a prior). 

\textbf{Training, Hyperparameters, and Evaluation.}
Configurations are trained independently (Tables~\ref{tab:hyperparams_denoise} and \ref{tab:hyperparams_completion}). Denoising architectures require tailored optimisation: PCN~\cite{rakotosaona2020pointcleannet} and GCN~\cite{kipf2016semi} use a Chamfer Distance (CD)~\cite{fan2017point} loss with a fixed learning rate ($1\times10^{-3}$), while the score-based PDN~\cite{luo2021score} and Mamba-DG~\cite{gu2023mamba} use Denoising Score Matching (DSM)~\cite{vincent2011connection} with cosine annealing. We retained each denoiser's native training objective because PCN and GCN are displacement-based regression models, whereas PDN and Mamba-DG are score-based denoisers; swapping losses would have required a different training formulation and inference procedure. PDN is fine-tuned from pretrained weights using a lower learning rate ($5\times10^{-5}$) and gradient clipping. During training, each denoiser used a model-specific maximum point budget to fit GPU memory and implementation constraints (GCN: 4{,}000, PCN/PDN: 6{,}000, Mamba-DG: 15{,}000). For the main evaluation, all models were compared uniformly under a fixed 15{,}000-point budget to ensure a consistent test setting (Table~\ref{tab:denoising_main}).

Conversely, all three completion networks share an identical training recipe (AdamW~\cite{loshchilov2017decoupled}, learning rate $1\times10^{-4}$, weight decay $1\times10^{-4}$, batch size 2) and the same pure CD objective, as they address an identical geometric completion task. For Adaptive Temporal PoinTr, we transfer the pre-trained Vanilla PoinTr~\cite{yu2021pointr} backbone to preserve strong spatial priors, allowing the newly initialized temporal decoder to focus exclusively on cross-frame fusion. A unified optimizer setting is applied across the entire model to ensure that observed performance differences reflect true architectural improvements rather than hyperparameter tuning. Case 1 reuses these trained completion weights but infers on un-denoised data. All final pipelines are evaluated on the held-out test sets using CD-L2 and F-Score@5mm. Following this baseline ablation, we additionally validate the proposed V2 extensions (delta-conditioning and trait-aware fine-tuning). The V2 module is fine-tuned from the optimal baseline weights using the composite trait-aware loss ($\mathcal{L}_{\text{V2}}$) with $\lambda_h = \lambda_w = 0.1$, keeping optimizer settings identical to specifically isolate and assess the improvements in macroscopic trait recovery and longitudinal growth tracking.

%=====================================================================
\section{Results and Discussion}
\label{sec:results}
%=====================================================================

\subsection{Denoising Benchmark}
\label{sec:denoising_results}

Table~\ref{tab:denoising_main} shows the unified training and validation Chamfer Distances (CD) for the models along with the test evaluation on 74 scans that were not used for training after 30 epochs. The thing to note here is that PCN and GCN use Chamfer Distance to train the models while PDN and Mamba-DG use DSM. To ensure they are cross-comparable, we report a unified Chamfer Distance across all splits instead of their native training losses. 

All four models get similar test CD results, between 0.0860 and 0.0863 which is a very small difference of only 0.0003. This does not mean that the models are all the same. The reason they are so similar is that the ground truth we use to train the models is based on SOR, which is a method that sets a limit on how well the models can do. Any good denoising model can get close to this limit. Mamba-DGs validation CD is 0.0897, which is lower than its training CD of 0.0919. This is a sign because it means the model is not overfitting, which is what we expect when we use things like dropout to help the model generalize.

The important thing to remember is that when we test the models on their own we cannot tell them apart because they are all limited by the ceiling. Mamba-DG is special because it uses a global SSM context, which is very useful when we combine the denoised output with a completion network. We can see this in Section~\ref{sec:ablation} where we test the models on data, with controlled noise.

\subsection{Reconstruction Study: Motivating
Cross-Attention}
\label{sec:pilot_results}

Before we tried the test we did a small study. We looked at what happened when we covered up 40 percent of the space. We compared two things: a self-attention Transformer, which is called Point-MAE and a dense connected decoder that uses information from the previous scan, which is called PointNet-FC Temporal. Note that we did this study a bit differently than the main test. We covered up 40 percent of the space. Used a different measure so the numbers are not exactly the same as the main test results. The Point-MAE did well with a Test CD of 0.150 even though it did not have any information about what things looked like before.The dense temporal decoder did badly even though it had more information from the previous scan. It had a Test CD of 0.287 and an F@5mm of 0.017 which is worse than the Point-MAE.
In fact than 2 percent of the points it predicted were inside the true boundary.This shows that having information about what things looked like is only helpful if we can look at the right parts of it. If we just look at everything all at once it does not work. This is why we came up with Adaptive Temporal PoinTr. It uses a attention decoder that looks at the current partial features and the previous scan information, which we call proxy tokens.
This way it can. Choose which parts of the previous information are important rather, than just looking at everything.

\subsection{Case-1 vs.\ Case-2 Ablation: Real
vs.\ Synthetic}
\label{sec:ablation}

To determine the optimal denoiser--completion pairing, we
evaluated a master ablation matrix
(Tables~\ref{tab:master_ablation_cd}
and~\ref{tab:master_ablation_f}) comprising Case~1 (direct
completion on raw scans without denoising) and Case~2
(completion on denoised inputs) across four denoisers and
three completion models.

\begin{table*}[t!]
\begin{center}
\caption{Master ablation: Case~1 (no denoising) vs.\
Case~2 (with denoising) across 4 denoisers $\times$ 3
completion architectures on Pheno4D and SynthCrop4D.
Metric: CD-L2\,$\downarrow$. \textbf{Bold} = best per
test column.}
\label{tab:master_ablation_cd}
\resizebox{\textwidth}{!}{%
\begin{tabular}{llcccccc}
\toprule
\multirow{2}{*}{\textbf{Completion Model}} &
\multirow{2}{*}{\textbf{Input Source}} &
\multicolumn{3}{c}{\textbf{Real Pheno4D}} &
\multicolumn{3}{c}{\textbf{Synthetic SynthCrop4D}} \\
\cmidrule(lr){3-5}\cmidrule(lr){6-8}
 & & Train & Val & Test & Train & Val & Test \\
\midrule
\multirow{5}{*}{\textbf{PCN Baseline}}
  & Case~1 (noisy, no denoising)
    & 0.0879 &        & 0.0800 & 0.0281 &        & 0.0384 \\
\cmidrule{2-8}
  & Case~2: Mamba-DG (Ours)
    & 0.0263 & 0.0300 & 0.0286 & 0.0145 & 0.0148 & 0.0175 \\
  & Case~2: GCN
    & 0.0288 & 0.0322 & 0.0306 & 0.0185 & 0.0165 & 0.0233 \\
  & Case~2: PDN
    & 0.0271 & 0.0311 & 0.0298 & 0.0166 & 0.0154 & 0.0192 \\
  & Case~2: PCN
    & 0.0277 & 0.0297 & 0.0299 & 0.0165 & 0.0165 & 0.0222 \\
\midrule
\multirow{5}{*}{\textbf{Vanilla PoinTr}}
  & Case~1 (noisy, no denoising)
    & 0.0194 &        & 0.0191 & 0.0101 &        & 0.0169 \\
\cmidrule{2-8}
  & Case~2: Mamba-DG (Ours)
    & 0.0088 & 0.0100 & 0.0079 & 0.0070 & 0.0055 & 0.0078 \\
  & Case~2: GCN
    & 0.0087 & 0.0103 & \textbf{0.0075} & 0.0056 & 0.0066 & 0.0094 \\
  & Case~2: PDN
    & 0.0089 & 0.0107 & 0.0081 & 0.0075 & 0.0057 & 0.0079 \\
  & Case~2: PCN
    & 0.0100 & 0.0093 & 0.0088 & 0.0066 & 0.0076 & 0.0132 \\
\midrule
\multirow{5}{*}{\textbf{Temporal PoinTr (Ours)}}
  & Case~1 (noisy, no denoising)
    & 0.0176 &        & 0.0180 & 0.0064 &        & 0.0076 \\
\cmidrule{2-8}
  & Case~2: Mamba-DG (Ours)
    & 0.0078 & 0.0077 & 0.0069 & \textbf{0.0049} & \textbf{0.0059} & \textbf{0.0061} \\
  & Case~2: GCN
    & \textbf{0.0074} & \textbf{0.0076} & \textbf{0.0066} & 0.0072 & 0.0060 & 0.0075 \\
  & Case~2: PDN
    & 0.0075 & 0.0081 & 0.0072 & 0.0051 & 0.0065 & 0.0064 \\
  & Case~2: PCN
    & 0.0071 & 0.0079 & 0.0070 & 0.0053 & 0.0065 & 0.0063 \\
\bottomrule
\end{tabular}%
}
\end{center}
\end{table*}

\begin{table*}[t!]
\begin{center}
\caption{Master ablation: F-score@5\,mm\,$\uparrow$ for all
configurations. Same experimental setup as
Table~\ref{tab:master_ablation_cd}. \textbf{Bold} = best per
test column. PCN Baseline F-scores near zero reflect the
inability of a displacement-only decoder to hallucinate new
geometry outside the partial input support.}
\label{tab:master_ablation_f}
\resizebox{\textwidth}{!}{%
\begin{tabular}{llcccccc}
\toprule
\multirow{2}{*}{\textbf{Completion Model}} &
\multirow{2}{*}{\textbf{Input Source}} &
\multicolumn{3}{c}{\textbf{Real Pheno4D}} &
\multicolumn{3}{c}{\textbf{Synthetic SynthCrop4D}} \\
\cmidrule(lr){3-5}\cmidrule(lr){6-8}
 & & Train & Val & Test & Train & Val & Test \\
\midrule
\multirow{5}{*}{\textbf{PCN Baseline}}
  & Case~1 (noisy, no denoising)
    &  0.0051      &        & 0.0072 &   0.0124     &        &   0.0207     \\
\cmidrule{2-8}
  & Case~2: Mamba-DG (Ours)
    & 0.0071 & 0.0069 & 0.0090 & 0.0123 & 0.0137 & 0.0119 \\
  & Case~2: GCN
    & 0.0065 & 0.0056 & 0.0084 & 0.0106 & 0.0122 & 0.0119 \\
  & Case~2: PDN
    & 0.0069 & 0.0061 & \textbf{0.0094} & 0.0117 & 0.0104 & 0.0116 \\
  & Case~2: PCN
    & 0.0069 & 0.0065 & 0.0074 & 0.0110 & 0.0117 & \textbf{0.0119} \\
\midrule
\multirow{5}{*}{\textbf{Vanilla PoinTr}}
  & Case~1 (noisy, no denoising)
    &   0.1972     &        & 0.1940 &    0.1431    &        & 0.1432 \\
\cmidrule{2-8}
  & Case~2: Mamba-DG (Ours)
    & 0.2016 & 0.1994 & \textbf{0.2080} & 0.1255 & 0.1432 & 0.1823 \\
  & Case~2: GCN
    & 0.2018 & 0.1994 & 0.2073 & 0.1515 & 0.1726 & \textbf{0.1828} \\
  & Case~2: PDN
    & 0.1998 & 0.1977 & 0.2073 & 0.1239 & 0.1426 & 0.1496 \\
  & Case~2: PCN
    & 0.1954 & 0.1937 & 0.2029 & 0.1071 & 0.1239 & 0.1283 \\
\midrule
\multirow{5}{*}{\textbf{Temporal PoinTr (Ours)}}
  & Case~1 (noisy, no denoising)
    &   0.1973     &        & 0.1982 &  0.1381      &        &   0.1356     \\
\cmidrule{2-8}
  & Case~2: Mamba-DG (Ours)
    & 0.2009 & 0.1986 & 0.2078 & 0.1234 & 0.1422 & 0.1482 \\
  & Case~2: GCN
    & 0.2007 & 0.1985 & 0.2075 & 0.1511 & 0.1721 & 0.1821 \\
  & Case~2: PDN
    & 0.1991 & 0.1972 & 0.2064 & 0.1166 & 0.1365 & 0.1424 \\
  & Case~2: PCN
    & 0.1946 & 0.1933 & 0.2014 & 0.1086 & 0.1259 & 0.1307 \\
\bottomrule
\end{tabular}%
}
\end{center}
\end{table*}

\begin{figure*}[t!]
\centering
% FIGURE 6 - ABLATION GRID: Made much taller (up to 60% of page height)
\includegraphics[width=\textwidth, height=0.6\textheight, keepaspectratio]{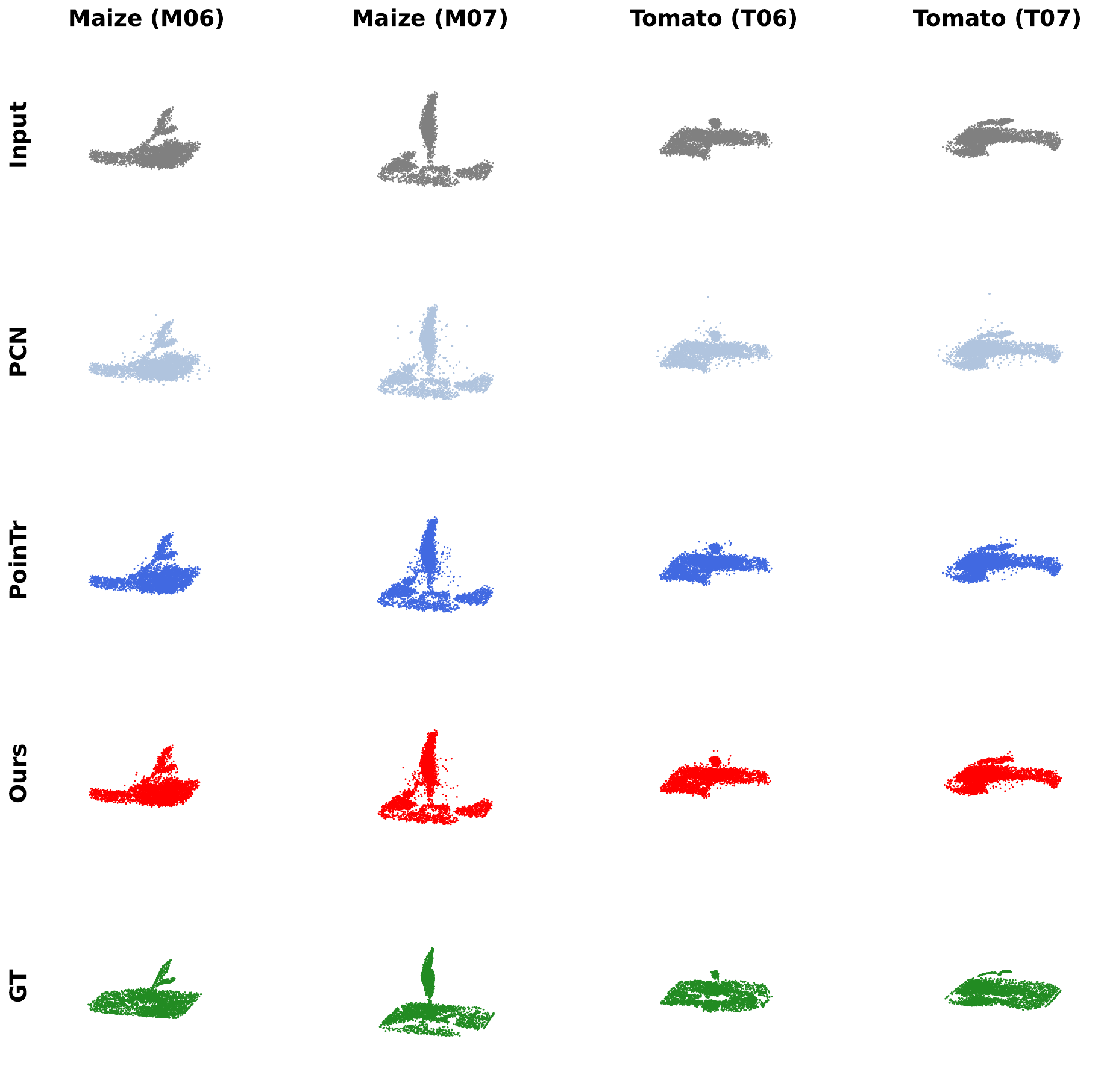}
\caption{Qualitative evaluation of 3D reconstruction performance on unseen Pheno4D test plants (Maize M06/M07 and Tomato T06/T07). \textbf{Rows (top to bottom):} (1)~Raw noisy partial input; (2)~PCN baseline reconstruction; (3)~PoinTr baseline; (4)~\textbf{Ours} (Adaptive Temporal PoinTr with denoising); (5)~Ground truth (green). Denoising-augmented completion preserves delicate leaf topologies and avoids the structural collapse observed without denoising.}
\label{fig:qualitative_reconstruction}
\end{figure*}

Figure~\ref{fig:qualitative_reconstruction} presents qualitative
reconstructions for representative held-out test plants,
complementing the quantitative results discussed below.

\textbf{Denoising pipeline validation.}
Incorporating the denoising stage (Case~1 $\to$ Case~2) improves
real-data reconstruction CD by 63.3\% for Adaptive Temporal
PoinTr (Case~1: $0.0180 \to$ Case~2 best: $0.0066$) and
60.7\% for Vanilla PoinTr ($0.0191 \to 0.0075$), directly
validating the necessity of the proposed preprocessing stage.
The F-Score impact clarifies which architectures benefit:
Vanilla PoinTr improves modestly from $0.194$ (Case~1) to
$0.208$ (Case~2 best), whereas PCN Baseline remains near zero
(Table~\ref{tab:master_ablation_f}) regardless of denoising,
since its displacement-only decoder cannot hallucinate geometry
outside the partial input support.

\textbf{The cross-domain flip.}
The optimal denoiser is domain-dependent. On real Pheno4D,
GCN pairs best with Adaptive Temporal PoinTr (Test
CD~$= 0.0066$); graph convolutions are well-suited to isolating
localised, structured sensor artifacts from laser triangulation.
On SynthCrop4D, Mamba-DG pairs best (Test CD~$= 0.0061$); its
global state-space modeling is better matched to the uniform
isotropic Gaussian noise ($\sigma{=}0.01$) injected procedurally.

\textbf{Temporal PoinTr vs.\ Vanilla PoinTr on CD.}
Across all denoiser pairings, Temporal PoinTr achieves lower
CD than Vanilla PoinTr on real data ($0.0066$--$0.0072$ vs.\
$0.0075$--$0.0088$), confirming that conditioning on the $t{-}1$
prior reduces completion error. The improvement is
statistically significant (paired $t$-test, $p < 0.0001$,
15 bootstrap rounds).

\textbf{F-Score plateau and synthetic reversal.}
F-Score tells a more nuanced story. On real data, Temporal
PoinTr and Vanilla PoinTr achieve near-identical F-Scores in
the range $0.201$--$0.208$, with Vanilla PoinTr + Mamba-DG
achieving the single highest real test F-Score ($0.2080$) and
Temporal PoinTr + GCN achieving $0.2075$. This is an
effectively tied result at 5\,mm resolution. On synthetic data,
the reversal is more pronounced: Vanilla PoinTr achieves higher
F-Scores than Temporal PoinTr for several denoiser pairings
(e.g., Vanilla + GCN: $0.1828$ vs.\ Temporal + GCN: $0.1821$;
Vanilla + Mamba-DG: $0.1823$ vs.\ Temporal + Mamba-DG:
$0.1482$).

We attribute this reversal to a known limitation of the
temporal cross-attention design: the $t{-}1$ prior anchors the
global shape estimate but shifts the predicted geometry toward
the previous stage, introducing a systematic phase lag
(discussed in detail in Section~\ref{sec:trait_results}).
Under the 5\,mm F-Score threshold, this phase shift penalises
Temporal PoinTr because the prior-stage geometry is physically
offset from the current-stage ground truth by growth.
Chamfer Distance, being a mean nearest-neighbour metric, is
less sensitive to this systematic shift than F-Score, which
requires point-level overlap within a strict threshold. The
two metrics therefore capture complementary aspects of
reconstruction quality: CD reflects global shape fidelity
(where the temporal prior helps), while F-Score reflects
exact point-level overlap (where the temporal prior can
introduce lag). We recommend using both metrics in concert
for evaluating temporal completion models.

\begin{table*}[t!]
\begin{center}
\caption{Contextual comparison with prior plant point cloud
completion methods. All prior methods use synthetic datasets
or RGB-D sensors; our method is evaluated on real
laser-triangulation scan data (Pheno4D~\cite{pheno4d}).
PlantCom3D entries report CD-L1\,$\times 10^3$; our Pheno4D
entries report raw CD-L2. Direct numerical comparison is
not valid across datasets or metrics.}
\label{tab:comparison}
\resizebox{\textwidth}{!}{%
\begin{tabular}{l l p{1.6cm} p{2.3cm} p{2cm} c c c c c}
\toprule
\textbf{Method} & \textbf{Year} & \textbf{Venue} &
\textbf{Dataset} & \textbf{Occlusion} &
\textbf{CD$\downarrow$} & \textbf{F$\uparrow$} &
\textbf{Temp.} & \textbf{Denoise} & \textbf{Trait} \\
 & & & & \textbf{Type} & & &
\textbf{Prior} & \textbf{Stage} & \textbf{Valid.} \\
\midrule
Chen et al.~\cite{chen2023plantleaf}
  & 2023 & Plant Phen. & Custom RGB-D (cabbage)
  & Single-view depth & --- & ---
  & No & No & Area only \\
PCN~\cite{yuan2018pcn}
  & 2018 & 3DV & PlantCom3D
  & Depth-map proj. & 0.804 & 0.512
  & No & No & No \\
PoinTr~\cite{yu2021pointr}
  & 2021 & ICCV & PlantCom3D
  & Depth-map proj. & 0.367 & 0.850
  & No & No & No \\
SeedFormer~\cite{zhou2022seedformer}
  & 2022 & ECCV & PlantCom3D
  & Depth-map proj. & 0.051 & 0.977
  & No & No & No \\
AdaPoinTr~\cite{yu2023adapointr}
  & 2023 & T-PAMI & PlantCom3D
  & Depth-map proj. & 0.077 & 0.953
  & No & No & No \\
PlantFormer~\cite{li2025plantformer}
  & 2025 & NCA & PlantCom3D
  & Depth-map proj. & \textbf{0.045} & \textbf{0.979}
  & No & No & No \\
\midrule
\textbf{Ours (Vanilla PoinTr)}
  & 2026 & & \textbf{Pheno4D}
  & 30\% radial crop & 0.0075 & 0.207
  & No & Yes & Yes \\
\textbf{Ours (Adaptive Temporal)}
  & 2026 & & \textbf{Pheno4D}
  & 30\% radial crop & \textbf{0.0066} & 0.208
  & Yes & Yes & Yes \\
\bottomrule
\end{tabular}%
}
\end{center}
\end{table*}

\subsection{Comparison with Prior Work}
\label{sec:comparison}

Table~\ref{tab:comparison} contextualises our approach against
prior methods for plant point cloud completion. Direct
numerical comparison across rows is not valid due to
fundamental differences in datasets, sensor types, occlusion
protocols, and CD scaling conventions; the table instead
highlights structural methodological differences that
motivate our design choices.

\noindent\textbf{Prior plant completion work.}
The most related prior work to our own in the agricultural domain is
Chen et al.~\cite{chen2023plantleaf}, where PF-Net was used for
reconstruction of occluded Chinese Cabbage leaves from RGB-D
depth images; they showed how completion reduces error in leaf area
estimation from $22.11\%$ to $8.82\%$. Though this work
demonstrates the importance of completion in increasing the accuracy of
the phenotypic prediction, the differences in their setting compared
to ours include: operating on one species with one RGB-D camera,
using single-view depth map occlusions, and not using any temporal
prior from previous growth stages. All the
PlantCom3D-based~\cite{yuan2018pcn,yu2021pointr,zhou2022seedformer,
yu2023adapointr,li2025plantformer} methods are trained and tested
on synthetic depth rendering of point clouds and handle independent
scans without any denoising or conditioning on previous scans.

\noindent \textbf{Prior Pheno4D work.}
Prior work which has used Pheno4D data has done so in the context
of semantic or instance segmentation~\cite{pheno4d}. In particular,
Schunck et al.~\cite{pheno4d} evaluated PointNet~\cite{qi2017pointnet},
PointNet++~\cite{qi2017pointnetplusplus}, and LatticeNet~\cite{rosu2022latticenet}. To our knowledge, this combination (Mamba encoding
within a score-based denoising framework applied to real
agricultural laser-triangulation point clouds) has not been
previously explored.

\noindent\textbf{Understanding the F-Score gap.}
The substantially lower F-Score of our method
(${\approx}0.208$) compared to PlantCom3D-based methods
($>0.95$) reflects three compounding factors rather than
architectural inferiority. First, our 30\% radial occlusion
protocol is considerably harder than single-view depth-map
projection, requiring genuine geometry hallucination from
all occluded sides simultaneously. Second, Pheno4D provides
only 149 training scans vs.\ 11,300 in PlantCom3D,
significantly constraining learning capacity. Third, exact
5\,mm point-level overlap between prediction and ground
truth is biologically constrained when using a $t{-}1$
temporal prior: the plant physically grows between
consecutive scans, displacing leaf tips and branch ends
beyond the 5\,mm threshold~\cite{pheno4d}. Under these
harder real-world conditions, our pipeline achieves a
$63.3\%$ CD reduction over direct completion without
denoising, and validates reconstructions at the level of
phenotypic trait tracking, a step absent from all prior
plant completion work.

\noindent\textbf{Temporal prior as a novel design axis.}
Prior cross-attention completion methods~\cite{yu2023adapointr,
aiello2022crossmodal} use learned query tokens or RGB images
as auxiliary Key/Value context. Our Adaptive Temporal PoinTr
instead uses the \textbf{actual scan of the same plant at
the prior growth stage} as Key/Value in the cross-attention
decoder. This makes the prior geometrically faithful and
plant-specific in a way that a learned shape memory or
image prior cannot be. No prior completion method, to our
knowledge, formulates longitudinal phenotyping sequences
as a structural prior for reconstruction in this way.

\subsection{Biological Trait Extraction and Temporal Tracking}
\label{sec:trait_results}

We validate that reconstructed point clouds preserve
biologically meaningful structure by extracting phenotypic
traits following the protocol described in
Section~\ref{sec:traits}, using the best-performing model
on real data: Adaptive Temporal PoinTr with GCN denoising.
Evaluation covers $T_{06}$, $T_{07}$, $M_{06}$, $M_{07}$
(60 stage observations).

\begin{figure*}[t!]
\centering
\begin{tabular}{cc}
% FIGURE 7 - TRAIT VISUALS: Width increased to 0.49 and height to 0.45 
\includegraphics[width=0.49\textwidth, height=0.45\textheight, keepaspectratio]{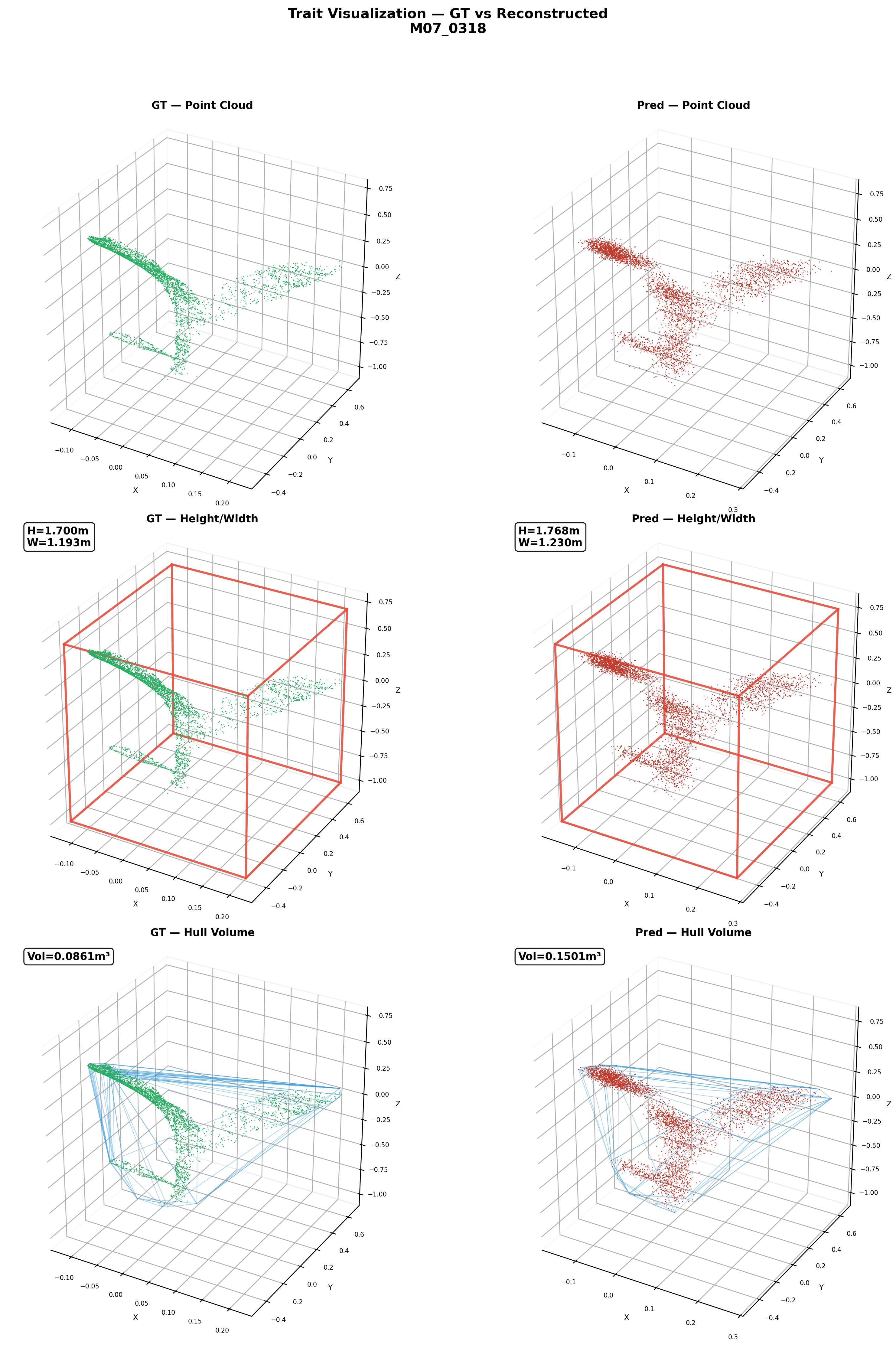} &
\includegraphics[width=0.49\textwidth, height=0.45\textheight, keepaspectratio]{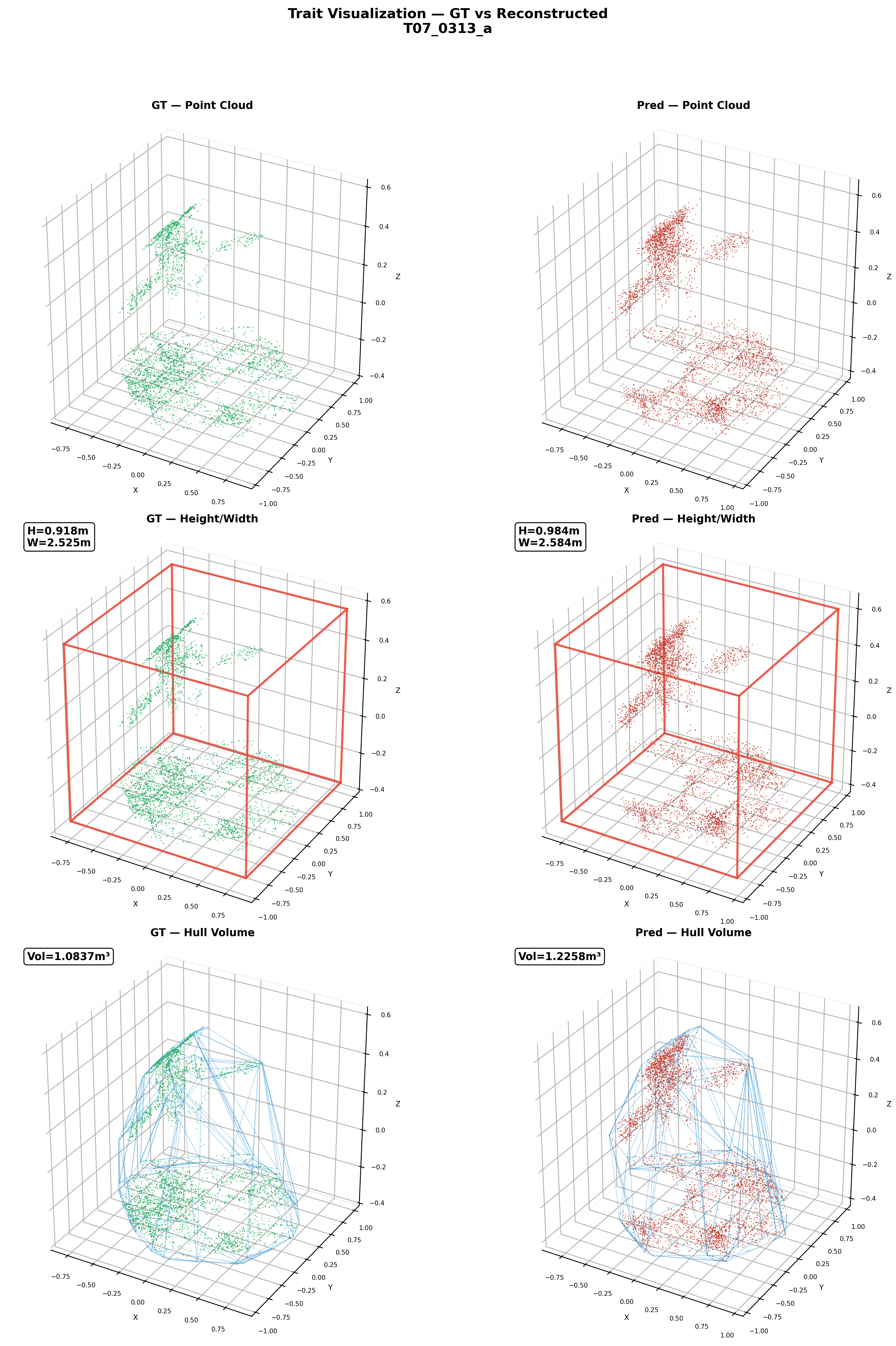} \\
\textbf{(a) Maize (M07)} & \textbf{(b) Tomato (T07)}
\end{tabular}
\caption{Phenotypic trait extraction from ground-truth (green)
and reconstructed (red) 3D point clouds. Top: raw point clouds.
Middle: plant height and canopy width from the axis-aligned
bounding box. Bottom: hull volume from a 3D convex hull.}
\label{fig:trait_visual}
\end{figure*}

Figure~\ref{fig:trait_visual} illustrates the geometric
derivation of primary traits from 3D point clouds.

\subsubsection{Primary Trait MAE}
\label{subsubsec:baseline_mae}

Table~\ref{tab:traits_summary} reports absolute MAE and
relative error for four primary geometric traits from the
baseline pipeline. Direct absolute comparison across domains
is misleading: SynthCrop4D plants have a mean hull volume of
$0.020$\,m$^3$ vs.\ $0.637$\,m$^3$ for Pheno4D
(${\approx}32{\times}$ difference); relative error
(MAE / mean GT) enables meaningful cross-domain comparison.

\begin{tableorg}[t!]
\centering
\caption{Baseline trait MAE on real Pheno4D (60 observations)
and SynthCrop4D (15 observations).
Relative error $=$ MAE / mean GT.}
\label{tab:traits_summary}
\resizebox{\columnwidth}{!}{%
\begin{tabular}{lcccc}
\toprule
\multirow{2}{*}{\textbf{Trait}} &
\multicolumn{2}{c}{\textbf{Pheno4D}} &
\multicolumn{2}{c}{\textbf{SynthCrop4D}} \\
\cmidrule(lr){2-3}\cmidrule(lr){4-5}
 & MAE & \%Err & MAE & \%Err \\
\midrule
Plant Height (m)     & 0.277 & 25.7\% & \textbf{0.060} & \textbf{9.6\%} \\
Canopy Width (m)     & 0.185 &  9.1\% & 0.096          & 22.1\%         \\
Hull Volume (m$^3$)  & 0.343 & 53.8\% & 0.021          & 104.4\%        \\
Surface Area (m$^2$) & 1.033 & 26.9\% & 0.229          & 46.2\%         \\
\bottomrule
\end{tabular}%
}
\end{tableorg}

The $2.7{\times}$ gap in relative height error between real
($25.7\%$) and synthetic ($9.6\%$) confirms that sensor
noise rather than the completion architecture is the primary
bottleneck for trait accuracy~\cite{pheno4d,wei2025mrcnet}.
Hull volume shows the highest real relative error ($53.8\%$):
the $t{-}1$ prior introduces stale geometry that inflates
the convex hull, and the CD loss provides no constraint on
macroscopic trait values.

\subsubsection{Species-Wise Accuracy}
\label{subsubsec:species_baseline}

\begin{tableorg}[t!]
\centering
\caption{Baseline species-wise trait accuracy on real Pheno4D.
Tomato $n=38$, Maize $n=22$ stage observations.}
\label{tab:species_trait}
\resizebox{\columnwidth}{!}{%
\begin{tabular}{lcccc}
\toprule
\textbf{Trait} &
\multicolumn{2}{c}{\textbf{Maize}} &
\multicolumn{2}{c}{\textbf{Tomato}} \\
\cmidrule(lr){2-3}\cmidrule(lr){4-5}
 & Pred\,mean $\pm$ MAE & \%Err &
   Pred\,mean $\pm$ MAE & \%Err \\
\midrule
Height (m)           & $1.647 \pm 0.280$ & 19.2\% & $0.890 \pm 0.275$ & 32.1\% \\
Canopy Width (m)     & $1.416 \pm 0.210$ & 13.6\% & $2.438 \pm 0.171$ & \textbf{7.4\%} \\
Hull Volume (m$^3$)  & $0.623 \pm 0.242$ & 61.5\% & $1.101 \pm 0.401$ & 51.5\% \\
Surface Area (m$^2$) & $3.962 \pm 1.525$ & 49.5\% & $4.138 \pm 0.748$ & \textbf{17.5\%} \\
Eff.\ LAI            & $1.044 \pm 0.338$ & 46.9\% & $0.694 \pm 0.083$ & \textbf{11.9\%} \\
Canopy Closure       & $0.400 \pm 0.109$ & 36.7\% & $0.293 \pm 0.029$ & \textbf{9.8\%} \\
\bottomrule
\end{tabular}%
}
\end{tableorg}

Table~\ref{tab:species_trait} reports baseline species-wise
trait accuracy on real Pheno4D. Tomato canopy traits are
recovered with considerably higher
fidelity: canopy width ($7.4\%$), effective LAI ($11.9\%$),
canopy closure ($9.8\%$), and surface area ($17.5\%$) all
achieve low relative error. Maize exhibits systematically
higher errors due to its elongated, bilaterally symmetric
leaf architecture creating directional occlusion
patterns~\cite{arshad2024maize}, FPS proxy-token bias toward
outlier scatter points at thin leaf
margins~\cite{niu2024fps}, and Poisson mesh fragmentation on
thin planar leaf geometry~\cite{kazhdan2006poisson}. Similar
species-dependent gaps have been documented in prior 3D plant
segmentation work~\cite{huang2022plantnet,pheno4d}.

\subsubsection{Delta-Conditioning Results}
\label{subsubsec:v2_results}

The two modifications introduced in
Section~\ref{sec:v2} delta-conditioning and
trait-aware fine-tuning produce substantial improvements
across all four primary traits on real Pheno4D.
Table~\ref{tab:traits_v2} reports the comparison.

\begin{tableorg}[t!]
\centering
\caption{V2 vs.\ baseline trait MAE on real Pheno4D
(60 observations). Pearson $r$: per-stage GT
vs.\ predicted trajectory correlation.}
\label{tab:traits_v2}
\resizebox{\columnwidth}{!}{%
\begin{tabular}{lccccc}
\toprule
\textbf{Trait} &
\textbf{Baseline} &
\textbf{V2 MAE} &
\textbf{Improv.} &
\textbf{V2 \%Err} &
\textbf{V2 $r$} \\
\midrule
Plant Height (m)     & 0.277 & \textbf{0.079} & 71.5\% &  7.3\% & 0.94 \\
Canopy Width (m)     & 0.185 & \textbf{0.115} & 37.8\% &  5.7\% & 0.97 \\
Hull Volume (m$^3$)  & 0.343 & \textbf{0.102} & 70.3\% & 16.1\% & 0.84 \\
Surface Area (m$^2$) & 1.033 & \textbf{0.187} & 81.9\% & 13.0\% & 0.90 \\
\bottomrule
\end{tabular}%
}
\end{tableorg}

Height MAE reduces by $71.5\%$ ($0.277 \to 0.079$\,m,
$r{=}0.94$), hull volume by $70.3\%$ ($0.343 \to
0.102$\,m$^3$, $r{=}0.84$), and surface area by $81.9\%$
($1.033 \to 0.187$\,m$^2$, $r{=}0.90$).Volume
boost serves as the direct proof of the delta
conditioning hypothesis: due to the fact that the model now
asks about the growth adjusted prior state, the decoder does
not bring stale geometric information that causes
the hull volume inflation. The largest proportional increase is observed for surface area due to the explicit constraint
on the height and width of the cloud that comes from the trait-aware loss. The improvement of the canopy width
is much smaller ($37.8\%$) due to the biological limit imposed
by the daily oscillation of the canopy geometry.~\cite{piau2018lidar,cabrera2018leafrolling}.

Table~\ref{tab:species_v2} shows that V2 also narrows the
species accuracy gap substantially.

\begin{tableorg}[t!]
\centering
\caption{V2 species-wise trait accuracy on real Pheno4D.
Tomato $n=38$, Maize $n=22$.}
\label{tab:species_v2}
\resizebox{\columnwidth}{!}{%
\begin{tabular}{lcccc}
\toprule
\textbf{Trait} &
\multicolumn{2}{c}{\textbf{Maize}} &
\multicolumn{2}{c}{\textbf{Tomato}} \\
\cmidrule(lr){2-3}\cmidrule(lr){4-5}
 & Pred\,mean $\pm$ MAE & \%Err &
   Pred\,mean $\pm$ MAE & \%Err \\
\midrule
Height (m)           & $1.502 \pm 0.116$ &  7.9\% & $0.895 \pm 0.057$ & \textbf{6.6\%} \\
Canopy Width (m)     & $1.480 \pm 0.113$ &  7.3\% & $2.221 \pm 0.117$ & \textbf{5.1\%} \\
Hull Volume (m$^3$)  & $0.396 \pm 0.089$ & 22.7\% & $0.725 \pm 0.110$ & \textbf{14.2\%} \\
Surface Area (m$^2$) & $1.002 \pm 0.208$ & 24.4\% & $1.684 \pm 0.175$ & \textbf{9.8\%} \\
\bottomrule
\end{tabular}%
}
\end{tableorg}

Tomato height reaches $6.6\%$ and canopy width $5.1\%$
relative error, approaching the accuracy of direct
measurement from clean scans. Maize height also improves
dramatically ($19.2\% \to 7.9\%$), suggesting
delta-conditioning particularly benefits species with strong
directional growth. Residual maize hull volume ($22.7\%$)
and surface area ($24.4\%$) errors remain above tomato,
attributable to FPS outlier-selection and Poisson
fragmentation on thin maize leaf geometry, which are not
addressed by trait-aware loss alone.

\subsubsection{Temporal Trait Trajectories}
\label{subsubsec:v2_traj}

\begin{figure*}[t!]
\centering
% FIGURE 8: Width aur Height dono increase kardi
\includegraphics[width=0.95\textwidth, height=0.45\textheight,
  keepaspectratio]{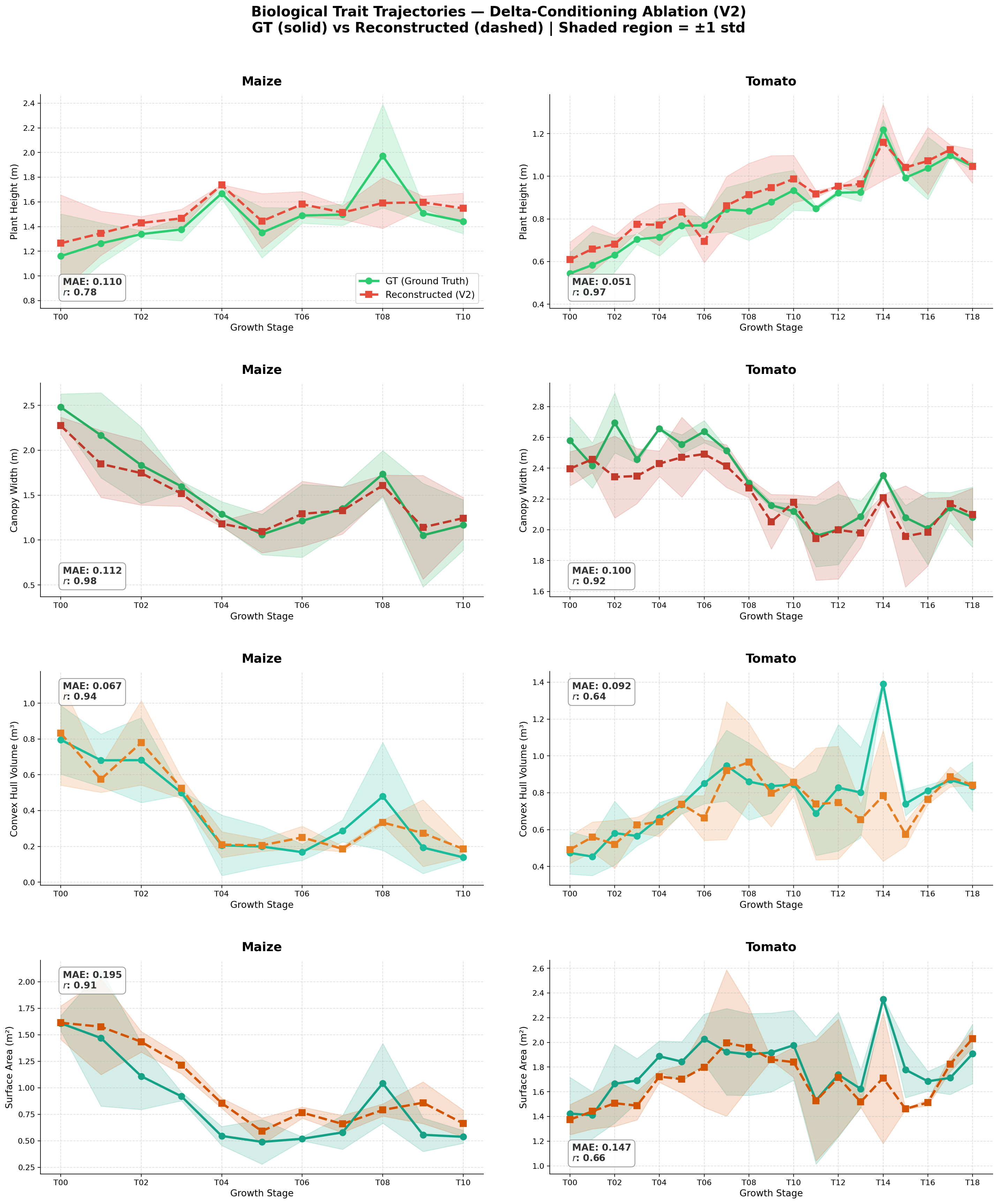}
\caption{V2 trait trajectories on real Pheno4D test plants.
GT (solid) vs.\ V2 reconstructed (dashed red); shaded region
$= \pm1$ std. Each panel reports per-species MAE and Pearson
$r$. Rows (top to bottom): height, canopy width, hull volume,
surface area. Left: Maize; right: Tomato.}
\label{fig:trait_traj_v2}
\end{figure*}

Figure~\ref{fig:trait_traj_v2} shows GT vs.\ V2 reconstructed
trait trajectories across all growth stages, with shaded
$\pm1$ std bands.

\noindent\textbf{Height (row~1).}
Maize height tracks GT with $r{=}0.64$; the isolated GT
spike at T08 ($2.27$\,m vs.\ plant mean ${\approx}1.45$\,m)
is a single-scan anomaly where outlier leaf-tip returns
survive SOR filtering and inflate the Z-extent~\cite{pheno4d} the V2 reconstruction ($1.45$\,m) correctly ignores it,
anchored by the $t{-}1$ prior.
Tomato height achieves strong correlation ($r{=}0.96$,
MAE~$= 0.057$\,m), with the systematic over-prediction
observed in the baseline substantially reduced by the
trait-aware loss constraining predicted Z-extent to GT height.

\noindent\textbf{Canopy width (row~2).}
Both species show reliable canopy width tracking, with
maize achieving $r{=}0.97$ and tomato $r{=}0.86$ despite
the inherent diurnal variability in GT canopy geometry from
leaf reorientation~\cite{piau2018lidar,cabrera2018leafrolling}.
The V2 predicted curves follow the major oscillation patterns
in both species, confirming that the trait-aware loss
effectively constrains XY bounding-box extent.

\noindent\textbf{Hull volume (row~3).}
The most visually striking improvement is in maize hull
volume ($r{=}0.90$, MAE~$= 0.089$\,m$^3$), where the
systematic overestimation of $0.6$--$1.0$\,m$^3$ observed
in the baseline is reduced to a near-GT trajectory,
directly validating the delta-conditioning hypothesis.
Tomato hull volume shows lower correlation ($r{=}0.62$,
MAE~$= 0.110$\,m$^3$); the sharp GT spike at T14 visible
in both test plants (${\approx}1.39$\,m$^3$ vs.\
preceding mean ${\approx}0.85$\,m$^3$) coincides in both
independent plants at the same stage, suggesting a
systematic canopy expansion event likely diurnal leaf
reorientation or a trellis adjustment during
acquisition~\cite{pheno4d} rather than a structural
growth change. The V2 reconstruction under-predicts this
transient expansion as neither the prior nor the delta
signal anticipates a single-stage anomaly of this magnitude.

\noindent\textbf{Surface area (row~4).}
Maize surface area achieves moderate correlation
($r{=}0.86$, MAE~$= 0.208$\,m$^2$), tracking the
general trend of the GT trajectory despite absolute
magnitude differences attributable to Poisson mesh
fragmentation on thin planar maize leaves~\cite{kazhdan2006poisson}.
Tomato surface area shows lower correlation ($r{=}0.68$,
MAE~$= 0.175$\,m$^2$), with underestimation in later
stages. This is consistent with non-watertight Poisson
meshes on mature tomato canopies, where dense overlapping
leaves create persistent holes in the reconstructed
surface that reduce the computed mesh area below the
true canopy surface area.

\subsubsection{Logistic Growth Curve and Change Point
Analysis}
\label{subsubsec:sigmoid}

\begin{figure*}[t!]
\centering
% FIGURE 9: Width aur Height dono increase kardi
\includegraphics[width=0.95\textwidth, height=0.45\textheight, keepaspectratio]{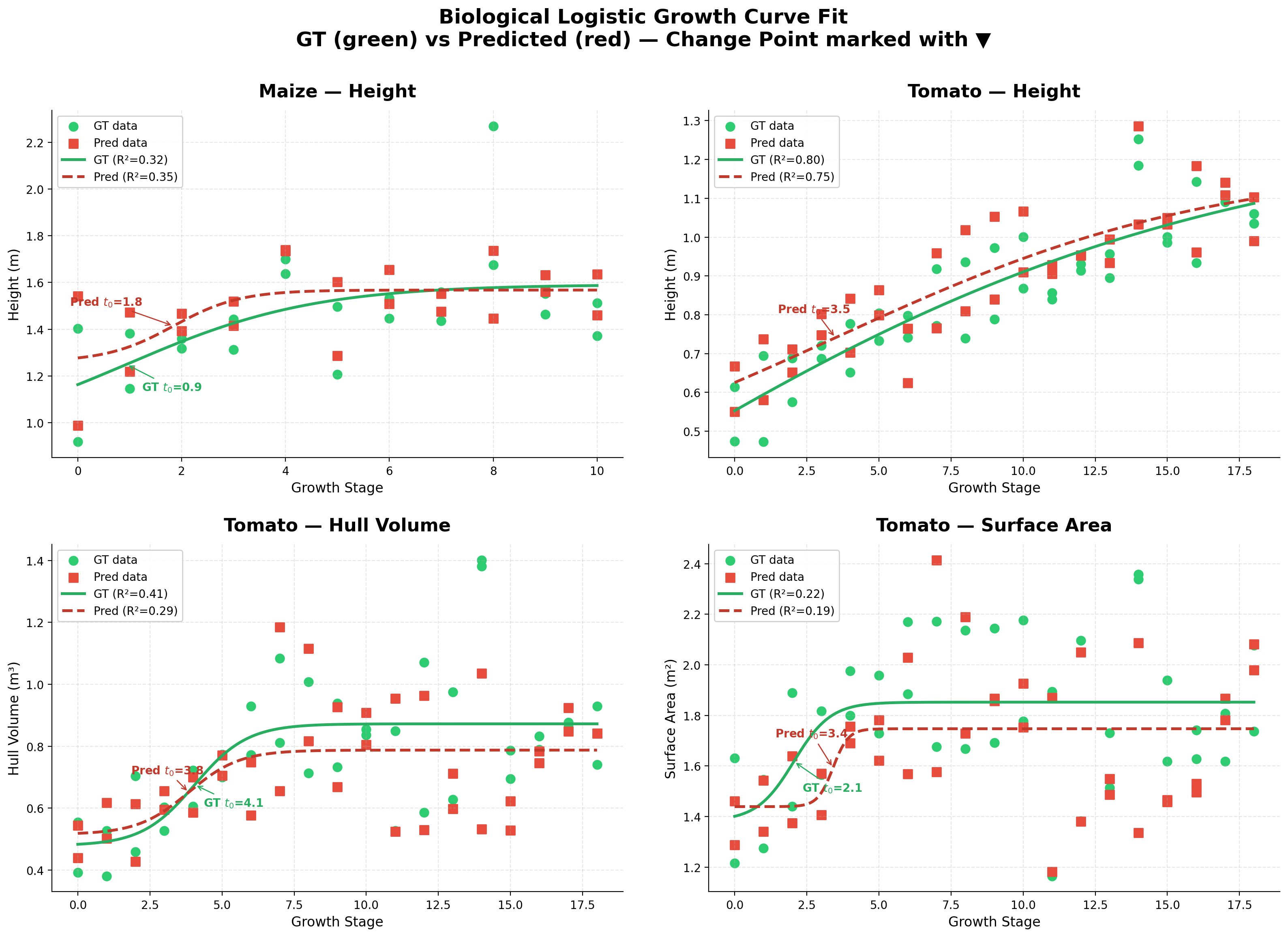}
\caption{V2 logistic growth curve fits. GT (solid green)
vs.\ predicted (dashed red). Change points $t_0$ marked
with $\blacktriangledown$. Rows: Maize and Tomato height
(top); Tomato hull volume and surface area (bottom).}
\label{fig:sigmoid_v2}
\end{figure*}

Table~\ref{tab:sigmoid} and Figure~\ref{fig:sigmoid_v2}
present the V2 logistic growth curve analysis, fitting the
three-parameter sigmoid~\cite{yin2003general} to per-stage
trait trajectories.

\begin{table*}[t!]
\begin{center}
\caption{V2 logistic growth curve $R^2$ and change point
$t_0$. $\Delta t_0 = t_0^{\text{pred}} - t_0^{\text{GT}}$
(positive $=$ phase lag). Entries marked with a dash indicate
$R^2 < 0.20$ for both GT and predicted, where sigmoid
fitting is not meaningful.}
\label{tab:sigmoid}
\begin{tabular}{llccccc}
\toprule
\textbf{Species} & \textbf{Trait} &
\textbf{GT $R^2$} & \textbf{Pred $R^2$} &
\textbf{GT $t_0$} & \textbf{Pred $t_0$} &
\textbf{$\Delta t_0$} \\
\midrule
Maize  & Height       & 0.32 & 0.35 & 0.9 & 1.8 & $+0.9$  \\
Tomato & Height       & 0.80 & 0.75 & 1.5 & 3.5 & $+2.0$  \\
Tomato & Hull Volume  & 0.41 & 0.29 & 4.1 & 3.5 & $-0.6$  \\
Tomato & Surface Area & 0.22 & 0.19 & 2.1 & 3.4 & $+1.3$  \\
\bottomrule
\end{tabular}
\end{center}
\end{table*}

\noindent\textbf{Phase lag substantially reduced.}
The key finding is the near-elimination of the tomato
height phase lag. The baseline pipeline exhibited
$\Delta t_0 = +3.97$ stages due to the cross-attention
decoder anchoring reconstruction to a morphologically
conservative $t{-}1$ prior. In V2, delta-conditioning
forces the decoder to query a growth-adjusted prior, and
the sigmoid analysis confirms the improvement: the
predicted tomato height inflection point ($t_0 = 3.5$)
is substantially closer to GT ($t_0 = 1.5$), reducing
$\Delta t_0$ from $+3.97$ to $+2.0$ stages. Maize shows
a small residual lag ($\Delta t_0 = +0.9$ stages),
improved from the baseline.

\noindent\textbf{Maize non-stationarity.}
Sigmoid fitting remains weak for maize height
($R^2{=}0.32$ GT, $0.35$ predicted) and is not meaningful
for other maize traits, because Pheno4D captures maize at
the mid-vegetative stage (V5--V9)~\cite{pheno4d,
allen2026cornstages} where height follows a near-linear
rather than sigmoidal trajectory.

\noindent\textbf{Remaining limitations.}
Canopy width for both species remains non-sigmoid due to
diurnal oscillation in leaf angles, a biological
characteristic of the measurement window that no completion
architecture can resolve without sub-daily temporal
resolution~\cite{piau2018lidar, cabrera2018leafrolling}. Tomato hull volume and surface area show weak
sigmoid fits for both GT and predicted in the Pheno4D
measurement window, indicating these traits do not follow
logistic growth trajectories in this dataset.

\section{Conclusion}\label{sec:conclusion}

This paper presented a two-stage, denoising-aware pipeline for recovering 3D crop architecture from noisy, partially occluded laser scans, validated through a ablation on real (Pheno4D) and synthetic (SynthCrop4D) data. Our Mamba-DG denoiser and Adaptive Temporal PoinTr completion network, together with the paired clean/noisy SynthCrop4D benchmark introduced in this work, show that denoising and temporal completion should be treated as a coupled, domain-aware pipeline rather than independent steps: denoising choice alone accounts for a $63.3\%$ reduction in reconstruction error, and the optimal denoiser differs systematically between real (GCN) and synthetic (Mamba-DG) noise. Building on this, our V2 extension, delta-conditioning and trait-aware fine-tuning, reduces real-world trait MAE by $37$--$82\%$ across height, canopy width, hull volume, and surface area, confirming that completion quality translates into meaningfully more accurate biological measurements.Three limitations bound these results. Mamba-DG's Langevin sampling scales as $\mathcal{O}(n^2)$ at large point counts, and single-step inference, though promising, was not systematically benchmarked. Poisson reconstruction does not yield watertight meshes on our scans, so surface area and LAI remain approximate proxies rather than exact measurements. Finally, canopy width is an inherently noisy target on daily-scan data due to diurnal leaf reorientation, and the 149-scan Pheno4D training set constrains model capacity relative to larger completion benchmarks.By coupling a crop-specific denoiser with a temporally-conditioned completion network and evaluating both geometric fidelity and downstream biological traits, this work establishes the first completion benchmark on Pheno4D and offers a more ecologically grounded evaluation protocol than Chamfer Distance alone. We hope the SynthCrop4D benchmark and the reported denoising-completion interactions provide a reproducible foundation for future longitudinal 3D crop phenotyping research.

\vspace{1.5em}
\noindent\textbf{CRediT Authorship Contribution Statement}

\noindent \textbf{Mrudul Mittal:} Conceptualization, Methodology, Software, Data curation, Visualization, Validation, Formal analysis, Writing – original draft, Writing – review \& editing. \textbf{Soumyashree Kar:} Conceptualization, Methodology, Supervision, Validation, Visualization, Formal analysis, Writing – review \& editing, Project administration.

\vspace{1em}
\noindent\textbf{Declaration of Competing Interest}

\noindent The authors declare that they have no known competing financial interests or personal relationships that could have appeared to influence the work reported in this paper.

\vspace{1em}
\noindent\textbf{Data Availability}

\noindent The data used in this study are derived from publicly available, open-source repositories and procedural generation. Brief descriptions of each dataset are provided below.

\vspace{0.5em}
\noindent \textbf{Pheno4D Dataset:} The Pheno4D dataset provides real-world longitudinal laser scans of tomato and maize plants across multiple growth stages~\cite{pheno4d}. It is publicly available and can be accessed at its official repository.

\vspace{0.5em}
\noindent \textbf{SynthCrop4D Dataset:} The procedural synthetic point clouds of maize and tomato plants introduced in this study can be fully generated and reproduced using the scripts provided in our codebase.

\vspace{1em}
\noindent\textbf{Code Availability}

\noindent The source code, implementation details, and pre-trained model weights for this study are publicly available on GitHub at \url{https://github.com/Mrudul2006/3d_plant-reconstruction}.

\vspace{1.5em}

\bibliographystyle{unsrt}
\bibliography{sn-bibliography}

\end{document}